\documentclass{article}
\usepackage{xcolor}
\usepackage[preprint]{corl_2026} % Use this for the initial submission.
\usepackage{graphicx}  
\usepackage{wrapfig}
\usepackage{graphicx}
\usepackage{subcaption}
\usepackage{xspace}
\usepackage{booktabs} 
\usepackage{algorithm}
\usepackage{wrapfig}
\usepackage{algpseudocode}
\usepackage{tabularx}
\usepackage{amsmath}
\usepackage{amsfonts}
\newcommand{\myenv}{BURGER\xspace}
\definecolor{orange}{rgb}{1,0.647,0}

\title{BURGER: \underline{B}enchmarking and \underline{U}nderstanding \\ \underline{R}obotic \underline{G}eneralization via \underline{E}mbodied d\underline{R}awing
\vspace{1em}
}
\author{
\begin{tabular}{c}
Xiangyu Xu$^{1,*}$ \quad
Bangshuai Peng$^{1,*}$ \quad
Ziyi Liu$^{1,*}$ \quad
Mengjie Zhao$^{1,*}$
\\
Chenhao Lu$^{1,2}$ \quad
Zhengyang Fan$^{1}$ \quad
Huazhe Xu$^{1,2,\dagger}$ \quad
Kaizhe Hu$^{1,2,\dagger}$
\\[0.6em]
$^{1}$Tsinghua University \quad $^{2}$Shanghai Qizhi Institute
\\[0.4em]
{\small $^{*}$Equal contribution. \quad $^{\dagger}$Corresponding authors.}
\end{tabular}
}

\begin{document}
\hypersetup{
  pdftitle={BURGER: Benchmarking and Understanding Robotic Generalization via Embodied dRawing},
  pdfauthor={Xiangyu Xu; Bangshuai Peng; Ziyi Liu; Mengjie Zhao; Chenhao Lu; Zhengyang Fan; Huazhe Xu; Kaizhe Hu},
  pdfsubject={Language-conditioned robotic drawing benchmark and data-generation pipeline}
}
\maketitle
\begin{abstract}

How effectively can vision-language-action (VLA) models learn diverse language-conditioned behaviors as training data scales?
Investigating this question requires large and varied robot datasets, whose collection on physical robots remains costly.
We introduce \myenv, a language-conditioned robotic drawing benchmark with a scalable, prompt-driven pipeline for generating continuous drawing trajectories, both in simulation and the real world.
In \myenv, an agent starts from the same blank canvas and controls a robot arm to draw the object or spatial composition specified by language, while each instruction allows an open-ended variety of valid drawings and stroke orders.
Using 28,200 simulated demonstrations spanning 282 drawing tasks, we evaluate five representative VLA models across data scales, language grounding, and multimodal behavior.
\myenv supports fine-grained analysis of language grounding, spatial composition, and action generation, revealing distinct failure patterns across VLA models.
Our experiments suggest that some VLA models can benefit from greater task diversity and demonstration scale, capture multimodal drawing behaviors, and generalize to unseen compositions of familiar objects.
Finally, we deploy a trained policy on a Franka Panda arm, demonstrating the ability to draw reactively under human intervention. Project homepage and code can be found \href{https://siyingzhao.github.io/BURGER/}{\textbf{here}}. 

\end{abstract}


\begin{figure*}[t]
    \centering
    \includegraphics[width=\textwidth]{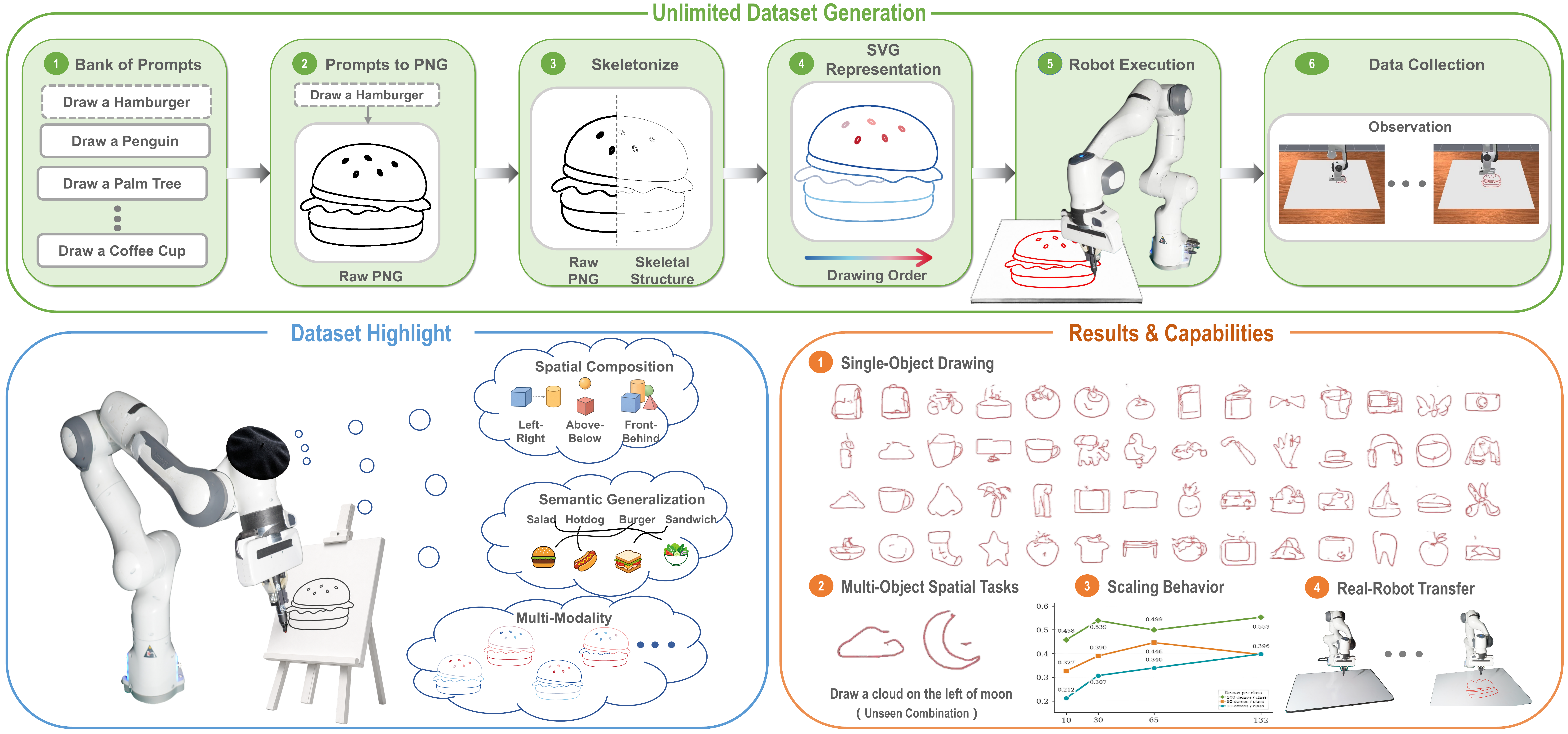}
    \caption{Overview of \myenv. 
\textbf{Top:} Prompt-generated PNG images are converted into skeletons, SVG strokes, and robot trajectories. 
\textbf{Bottom left:} \myenv~tests semantic grounding, spatial reasoning, and multimodal action generation in VLA models. 
\textbf{Bottom right:} We evaluate models on single-object drawing, compositional drawing, data scaling, and real-world deployment.
    }
    \label{fig:teaser}
    \vspace{-15pt}
\end{figure*}

\section{Introduction}
\label{sec:introduction}

Vision-language-action (VLA) models  integrate visual perception, language understanding, and action generation into policies for robotic control ~\cite{zitkovich2023rt,kim2024openvla,black2026pi0visionlanguageactionflowmodel,black2025pi05}.
As VLA models increasingly benefit from larger and more diverse datasets, an important question is whether current architectures are ready to effectively exploit large-scale language-conditioned robot trajectories.
Investigating this question requires robot datasets with controllable scale and diversity, together with evaluations that distinguish different aspects of model behavior~\cite{mees2022calvin,zhang2025vlabench,wang2025vlatest}.
However, collecting diverse demonstrations on physical robots remains costly~\cite{khazatsky2024droid,o2024open}, which motivates a more scalable setting for such studies.

Robotic drawing provides a complementary data regime with several practical advantages for such studies.
Starting from the same blank canvas, the agent cannot infer the requested drawing from the initial observation alone, eliminating a potential visual shortcut~\cite{xing2025shortcut}.
Language prompts can be systematically increased in semantic complexity by composing objects, attributes, and spatial relations.
The same instruction can admit many valid trajectories that differ in stroke order, starting point, and drawing direction, this intrinsic multimodality forces model to understand the concept, rather than memorizing the action sequence.
The resulting canvases and execution trajectories make semantic, spatial, and action-execution failures directly inspectable, supporting fine-grained diagnosis of model behavior.
Drawing therefore provides a scalable setting for testing whether VLA architectures can learn complex language-conditioned and multimodal robot trajectories before scaling up costly real-world data collection.

\begin{figure}[t]
    \centering
    \includegraphics[width=\columnwidth]{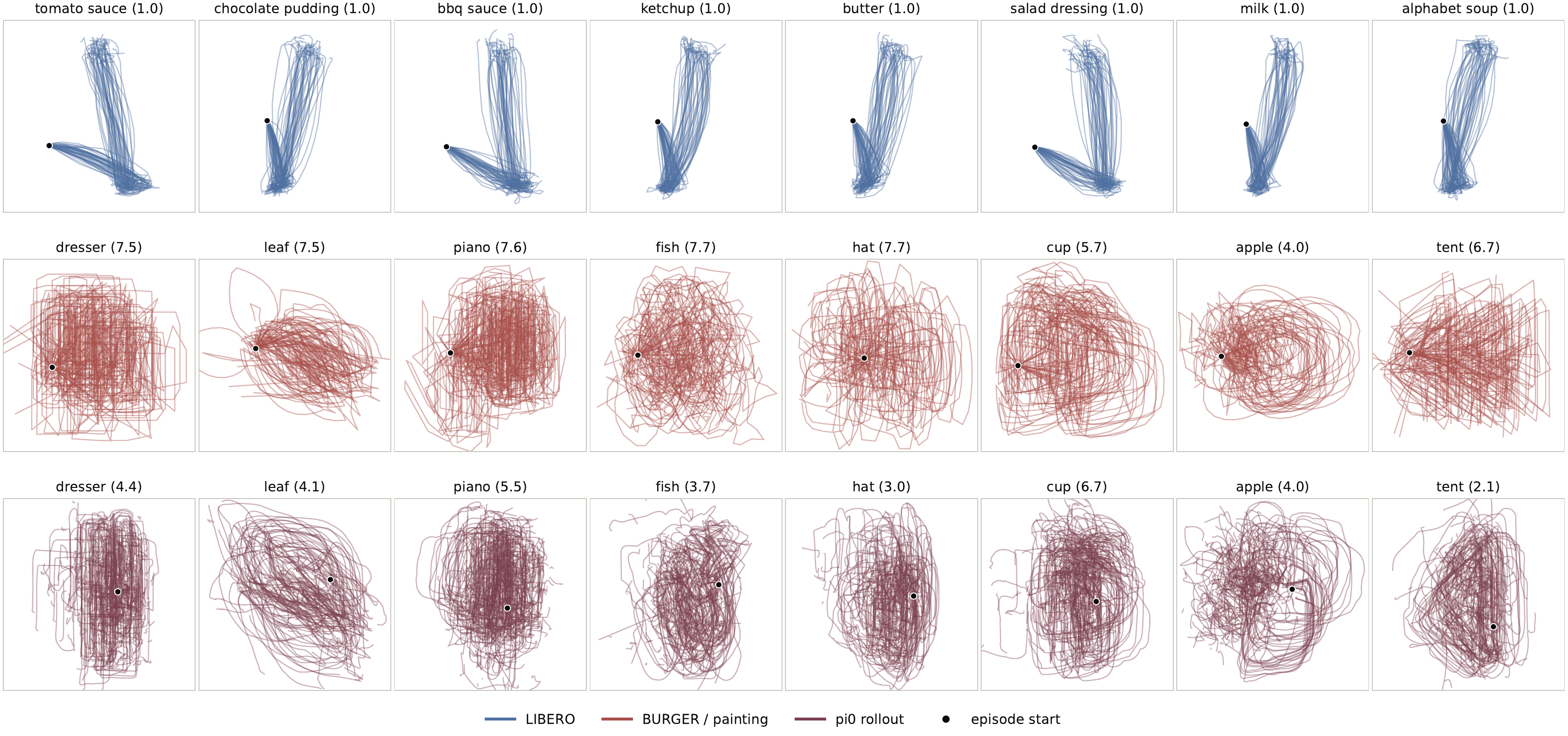}
    \caption{Instruction-conditioned planar trajectory diversity. The three rows show LIBERO demonstrations, BURGER demonstrations, and $\pi_{0}$ rollouts, respectively. Each subplot overlays trajectories for the same instruction. Different modes correspond to distinct clusters of execution routes.}
    \label{fig:data_comparison}
    \vspace{-10pt}
\end{figure}

Motivated by these properties, we introduce \myenv, a scalable benchmark and data-generation pipeline for language-conditioned robotic drawing, shown in Fig.~\ref{fig:teaser}.
Our pipeline generates target sketches from natural-language prompts and converts them into robot end-effector trajectories executable in simulation and on a real robot.
Using this pipeline, we construct 132 single-object tasks and 150 compositional tasks, with 100 demonstrations per task.
The dataset spans diverse objects, spatial compositions, and trajectory realizations.
As illustrated in Fig.~\ref{fig:data_comparison}, drawing trajectories in \myenv~exhibit substantially greater diversity in the plane for a given instruction than manipulation trajectories in LIBERO~\cite{liu2023libero}.
We introduce a DINOv3-based metric for scalable evaluation of generated drawings and validate its agreement with human judgments in a single-object user study.

We use \myenv~to examine how representative VLA models learn from larger datasets, capture trajectory diversity, and generalize to new compositions.
The flow-matching models generally improve as either the number of demonstrations per task or the number of training tasks increases, suggesting that some existing approaches can benefit from scaling robotic drawing data.
The $\pi$-family models~\cite{black2026pi0visionlanguageactionflowmodel, black2025pi05} can autonomously produce recognizable drawings after fine-tuning.
Beyond visual recognizability, $\pi_{0}$ retains multiple modes of the drawing trajectory distribution, as shown in Fig.~\ref{fig:data_comparison}.
We further observe that evaluated policies can generate unseen spatial compositions of familiar objects, but struggle to draw object categories absent from training.
Inspecting the resulting drawings and trajectories reveals semantic, spatial, and geometric failure patterns that are difficult to distinguish through task-success metrics alone.
We further demonstrate successful real-robot drawing after fine-tuning on 100 real-robot demonstrations alone, as well as sim-to-real transfer of a policy co-trained on 13,200 simulated \myenv~demonstrations.

Although robotic drawing simplifies the contact-rich interaction and object dynamics involved in manipulation, it retains important demands on language grounding and action generation.
Following a drawing instruction requires translating object descriptions and spatial relations into long-horizon continuous actions, while the range of valid drawings and stroke sequences provides a setting for evaluating multimodal behavior modeling.
These capabilities are also central to language-conditioned manipulation, making \myenv~a distinctive lens for examining the readiness and limitations of current VLA architectures in learning from larger and more diverse training datasets.

\section{Related Work}
\label{sec:related_work}

\textbf{Vision-language-action (VLA) models} map visual observations and language
instructions to robot actions. Recent VLAs broadly follow autoregressive or continuous
generative paradigms. Autoregressive models, such as OpenVLA~\cite{kim2024openvla},
predict actions sequentially, with recent variants improving action representations,
chunking, and parallel decoding~\cite{kim2025fine, pertsch2025fast, song2025pdvla,
wang2025vqvla, liu2025faster}. Continuous generative models produce action chunks
using flow matching or diffusion~\cite{lipman2023flowmatching}: $\pi_0$ and
$\pi_{0.5}$ use flow matching~\cite{black2026pi0visionlanguageactionflowmodel,
black2025pi05}, while RDT-1B and GR00T N1 adopt diffusion-based generation
\cite{liu2025rdt, bjorck2025gr00tn1}. Hybrid methods such as
DiffusionVLA~\cite{wen2025diffusionvla} combine autoregressive reasoning with
diffusion-based action generation. \myenv~evaluates these paradigms under
multimodal, long-horizon, and semantically rich drawing data.

\textbf{Benchmarking of VLA systems} has moved beyond measuring binary task success rates toward more diagnostic evaluation. Recent benchmarks evaluate VLA abilities in long-horizon reasoning, spatial and semantic grounding, robustness to scene or language perturbations, structured behavioral metrics, and memory-dependent manipulation~\cite{zhang2025vlabench, wang2025vlatest, wang2025roboeval, dai2026robomme, chen2026rmbench}. These efforts show that similar success rates can hide different failure modes in grounding, planning, robustness, or temporal consistency. \myenv~ follows this capability-oriented evaluation direction, but uses language-conditioned drawing as the diagnostic domain. Unlike goal-oriented manipulation, drawing admits many valid stroke orders, geometries, and trajectories for the same prompt. This makes the more dedicated abilities directly observable from the generated canvas and rollout trajectory.

\textbf{Data scaling in VLA learning} has been studied in robot imitation learning and generalist policy training. Prior work analyzes how data quantity and data composition affect manipulation-policy generalization, including object and environment diversity, camera viewpoints, spatial arrangements, and dataset fragmentation \cite{lin2025datascaling, saxena2025whatmatters, xing2025shortcut}. \myenv~ considers a related scaling question in a controlled language-conditioned drawing domain. % The benchmark varies the number of object classes, the number of trajectories per class, and the composition of prompts under a fixed task interface and evaluation pipeline. This design supports analysis of how semantic coverage and per-class trajectory coverage are associated with object recognizability, spatial composition, rollout diversity, and long-horizon stroke generation.

\textbf{Robotic drawing and painting} have been studied primarily as a control and planning problem. 
For example, Lee et al.~\cite{lee2022scratch} formulate robotic sketching as a deep decoupled hierarchical reinforcement learning problem, where separate policies learn stroke-based rendering and motor control from scratch. 
FRIDA~\cite{schaldenbrand2022frida} develops a collaborative robot painting system with a differentiable real2sim2real planning environment, enabling a robot to generate paintings from high-level human inputs such as language descriptions or reference images. 
More recently, IMPASTO~\cite{wang2026impasto} studies robotic oil-painting reproduction by integrating learned pixel dynamics models with model-based planning. 
These works advance robotic drawing by improving stroke planning, motor control, and real-world execution. 
In contrast, our work studies sketching as a language-conditioned VLA benchmark. 
% To our knowledge, \myenv~is the first large-scale robotic sketch-painting benchmark designed specifically for evaluating VLA models.
\section{Method}
\label{sec:method}

\begin{figure*}[t]
    \centering
    \includegraphics[width=\textwidth]{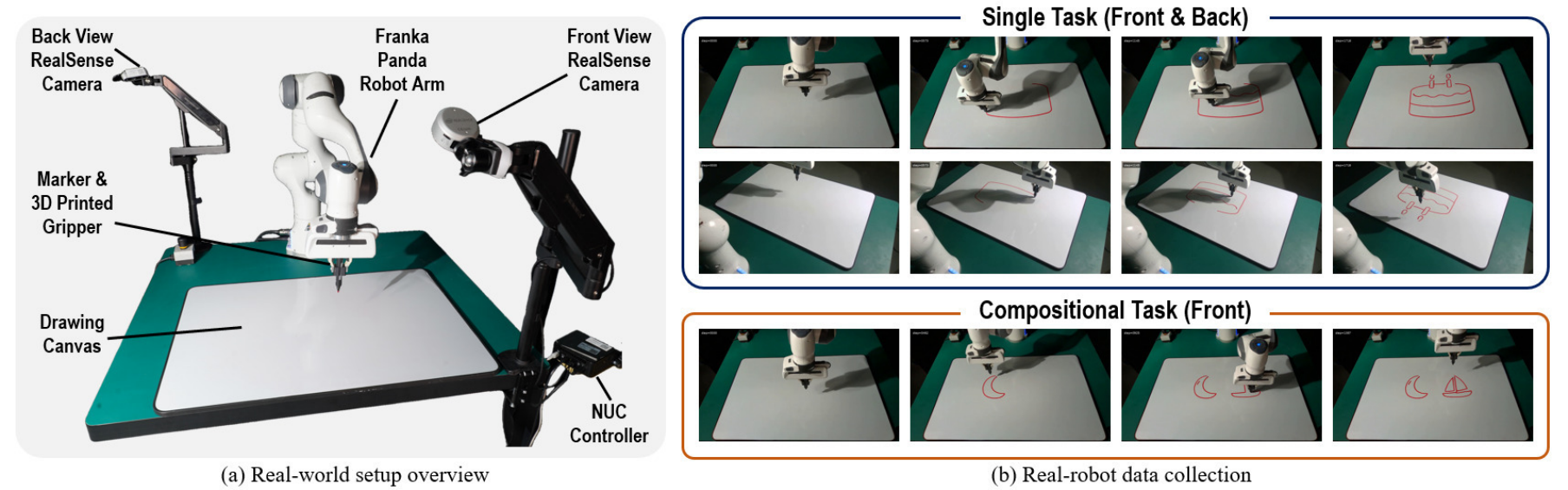}
    \caption{Real-world deployment of \myenv. \textbf{Left:} The Franka Panda hardware setup equipped with a drawing tool. \textbf{Right:} The real-robot data collection process for a single task ``birthday cake'' and a compositional task ``moon left sailboat''. }
    \label{fig:real_robot}

    \vspace{-10pt}
\end{figure*}

\subsection{Task Formulation}
\label{sec:task}
At time $t$, a policy receives a language instruction $\ell$, a canvas observation $I_t$, and robot proprioception $q_t$, and predicts an end-effector command $a_t$ (or an action chunk, depending on the model). The task is to produce the requested object or spatial composition through sequential drawing actions. The policy is not given a target sketch at inference. Canvas observations change as strokes accumulate, while pen-lifting transitions separate disconnected strokes. The benchmark therefore involves both drawing geometry and decisions about when to move without marking the canvas.

\subsection{Prompt-to-Trajectory Data Generation}

\begin{wrapfigure}[11]{r}{0.48\textwidth}
    \centering
    \vspace{-8pt}
    \includegraphics[width=\linewidth]{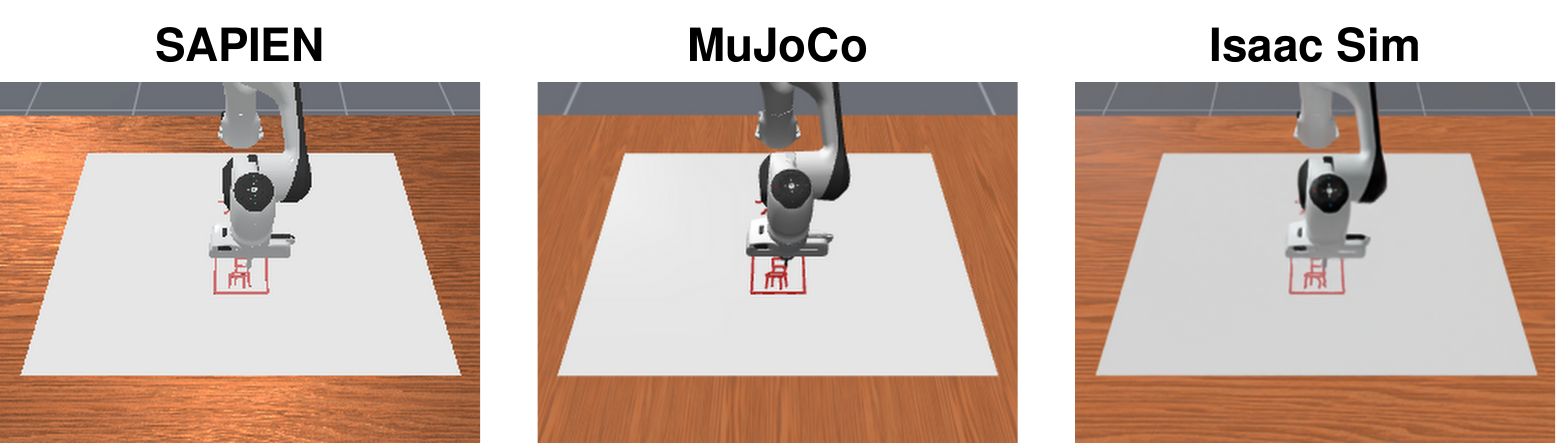}
    \caption{Cross-simulator deployment of the BURGER trajectory-generation pipeline. The same SVG drawing task is executed in SAPIEN, MuJoCo, and Isaac Sim using corresponding robot-control interfaces.}
    \label{fig:simulators}
    \vspace{-10pt}
\end{wrapfigure}

\label{sec:main_paper_pipline}
Figure~\ref{fig:teaser} summarizes our prompt-to-trajectory pipeline. Diffusion-based image generation models have shown
strong capabilities in producing diverse visual content from text prompts~\cite{zhang2023text}. We use GPT-Image 2~\cite{openai_gpt_image_2} to generate sketch candidates and convert them into ordered SVG strokes; Appendix~\ref{app:algorithms} describes the image-processing, skeletonization, and stroke-ordering steps. We replay each SVG in ManiSkill~\cite{taomaniskill3} using an end-effector pose controller, lifting the pen between disconnected strokes. We record RGB observations, robot proprioception, and actions as observation-action pairs, then convert the trajectories to the robot-dataset format. Although our main experiments
use ManiSkill with the SAPIEN backend, we also implement the pipeline in MuJoCo and Isaac Sim, as shown in Figure~\ref{fig:simulators}.

\subsection{The \myenv~Dataset}
\label{sec:dataset}

To collect representative and diverse drawing trajectories, we adopt the setting
from the Google Quick, Draw!~\cite{david2017neural} dataset, which contains 345
object classes and millions of human-drawn sketches. We select 132 classes that
are commonly found in the real world and can be easily drawn by a robot arm. We further define 150 compositional classes by pairing two of the 100 base classes
with spatial relations. Among them, 100 classes use left-right relations, while the
remaining 50 use vertical relations such as ``over'' and ``under''. For each class,
we generate 100 trajectories using the pipeline described above, resulting in 15,000
compositional trajectories and 28,200 trajectories in total.

To quantitatively characterize the advantage of \myenv~in providing multimodal
trajectory data, we compare its instruction-conditioned trajectory diversity with LIBERO, as visualized in Figure~\ref{fig:data_comparison}. Since
\myenv~focuses on drawing rather than full-space manipulation, we cluster planar
end-effector trajectories on the drawing plane with a shared codebook and estimate the effective
number of modes from the cluster frequencies. We use $k=24$ for the reported results and observe consistently higher multimodality for
\myenv~across different choices of $k$. Under this setting, LIBERO contains $1.09$
effective modes per instruction on average, compared with $7.13$ for \myenv,
corresponding to a $6.6\times$ increase. The same trajectory representation and clustering procedure is also used to
analyze policy rollouts in Sec.~\ref{sec:multimodal_behaviors}, and more details are provided in Appendix~\ref{app:dataset_comparison}.

We also collect real-robot trajectories to evaluate whether the BURGER data pipeline
can support real-world policy learning. The real-robot dataset follows the same data
format as the simulated dataset and includes two RealSense camera views from the front
and back of the workspace; either view can be used as the visual observation input.
Figure~\ref{fig:real_robot} shows the hardware setup and real-robot data-collection
process. The same pipeline supports both single-object and compositional drawing tasks,
as illustrated by the ``birthday cake'' and ``moon left of sailboat'' examples.
In this work, we collect 10 single-object classes with 10 demonstrations per class
for real-robot training and evaluation.

The dataset is, to our
knowledge, the first large-scale VLA dataset for language-conditioned drawing,
and we hope it can serve as a valuable resource for future research on VLA model
generalization. 
\begin{figure*}[t]
    \centering
    \includegraphics[width=\textwidth]{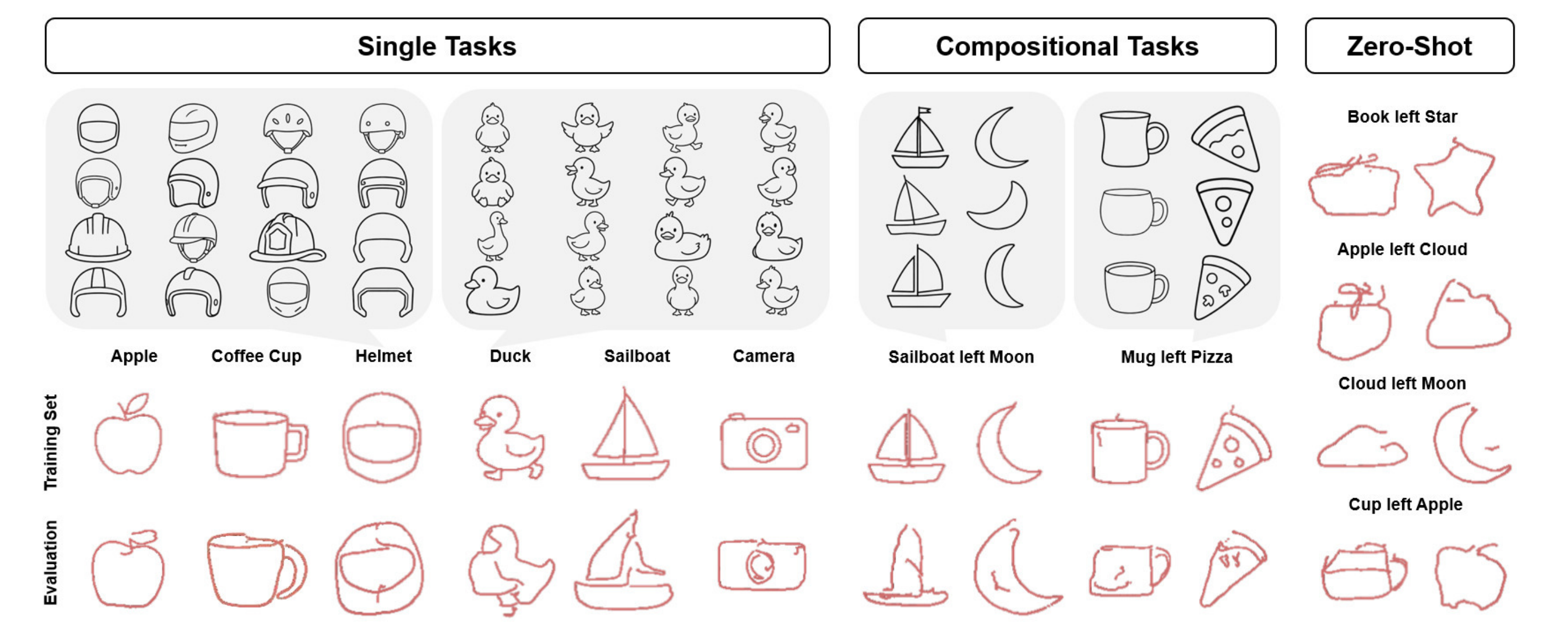}
    \caption{Dataset diversity and VLA evaluation in \myenv. We show representative
    single-object and compositional training tasks, along with corresponding VLA
    rollouts. The rightmost column evaluates generalization to unseen spatial compositions.
    }
    \label{fig:l}

\end{figure*}

\subsection{Quantifying Robot Trajectory Quality}
\label{sec:human_aligned_eval}

Valid drawings can differ in geometry and stroke order, making drawing quality difficult to evaluate with image-text similarity alone. Contrastive metrics such as CLIPScore~\cite{hessel2021clipscore}
measure global image--text alignment, but prior work~\cite{ahmadi2023robustness}
shows that such metrics can fail to detect fine-grained errors. 

To obtain a more human-aligned and fine-grained evaluation metric, we design a contrastive
feature-based evaluator trained directly on the \myenv~drawing distribution. Our evaluator builds a sketch-specific metric space on top of a frozen DINOv3 backbone~\cite{simeoni2025dinov3}, which allows us to score each drawing by its similarity to the target-class prototype. For compositional tasks, we separate the two objects at the minimum-connectivity boundary and evaluate their identities and spatial relation. Appendix~\ref{app:evaluation_details} provides the metric architecture, training objectives, and scoring procedure. In a user study covering 10 classes, our metric agrees more closely with human judgments than CLIPScore across Spearman correlation, Kendall's $\tau_b$, and pairwise agreement. Details are also provided in Appendix~\ref{app:evaluation_details}.

\section{Experimental Results}
\label{sec:result} Our experiments evaluate the capabilities and limitations of
VLA models on language-conditioned robotic drawing tasks in \myenv. Concretely,
we investigate three questions: (1) whether \myenv can serve as a diagnostic benchmark
for comparing VLA models across several dimensions, including data scaling, language
grounding, spatial-relation understanding, and multimodal behavior modeling; (2) whether
the benchmark can expose characteristic failure behaviors and help diagnose differences
among VLA formulations; and (3) whether policies
trained with BURGER data can transfer to real-world drawing and support closed-loop
interaction under human modifications.

\textbf{Baselines.} We evaluate representative VLA models with different
action-generation formulations. \textbf{$\boldsymbol{\pi}_{0}$} \cite{black2026pi0visionlanguageactionflowmodel}
uses a separate action expert to generate continuous actions using flow matching.
\textbf{$\boldsymbol{\pi}_{0.5}$} \cite{black2025pi05} extends $\pi_{0}$ with
co-training on heterogeneous tasks for broader generalization. \textbf{$\boldsymbol
{\pi}_{0}$-FAST} is an autoregressive VLA based on the FAST action tokenizer \cite{pertsch2025fast}.
\textbf{OpenVLA} \cite{kim2024openvla} is an open-source autoregressive VLA model.
\textbf{OpenVLA-OFT} \cite{kim2025fine} is an optimized fine-tuning variant of OpenVLA
that predicts continuous action chunks with parallel decoding, rather than predicting step-wise
discrete action tokens.

\subsection{Scaling Behavior of VLA Models}
\begin{figure}[hb]
    \centering
    \begin{subfigure}
        [t]{0.32\linewidth}
        \centering
        \raisebox{0\height}{\includegraphics[width=\linewidth]{
            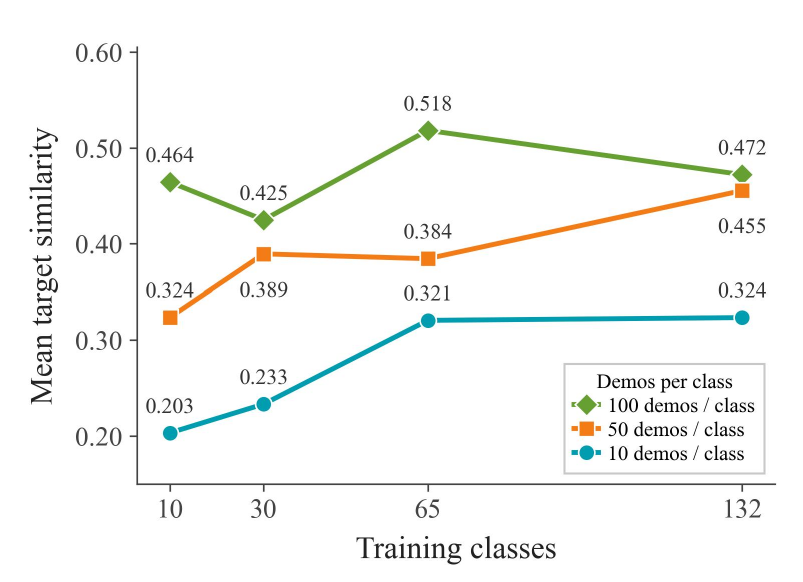
        }}
        \vspace{-15pt}
        \caption{$\pi_{0}$}
        \label{fig:pi0_scale}
    \end{subfigure}
    \begin{subfigure}
        [t]{0.32\linewidth}
        \centering
        \raisebox{0\height}{\includegraphics[width=\linewidth]{
            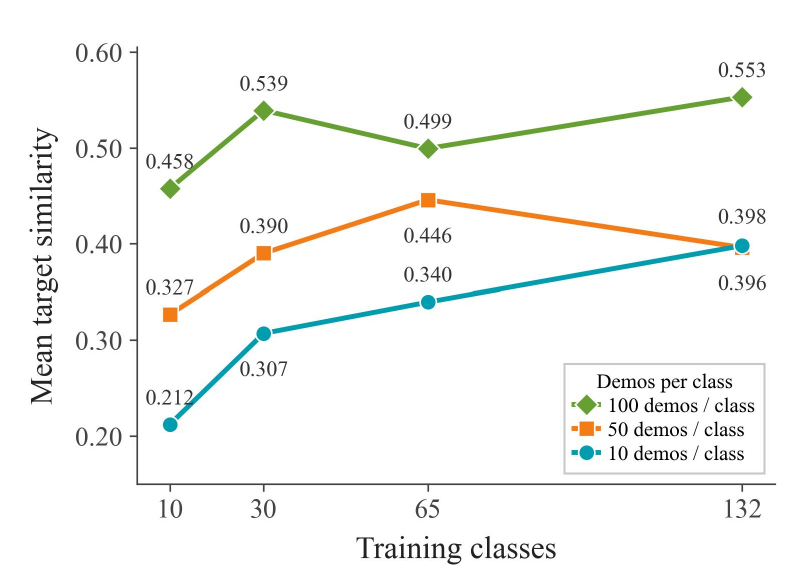
        }}
        \vspace{-15pt}
        \caption{$\pi_{0.5}$}
        \label{fig:pi05_scale}
    \end{subfigure}
    \begin{subfigure}
        [t]{0.32\linewidth}
        \centering
        \raisebox{0\height}{\includegraphics[width=\linewidth]{
            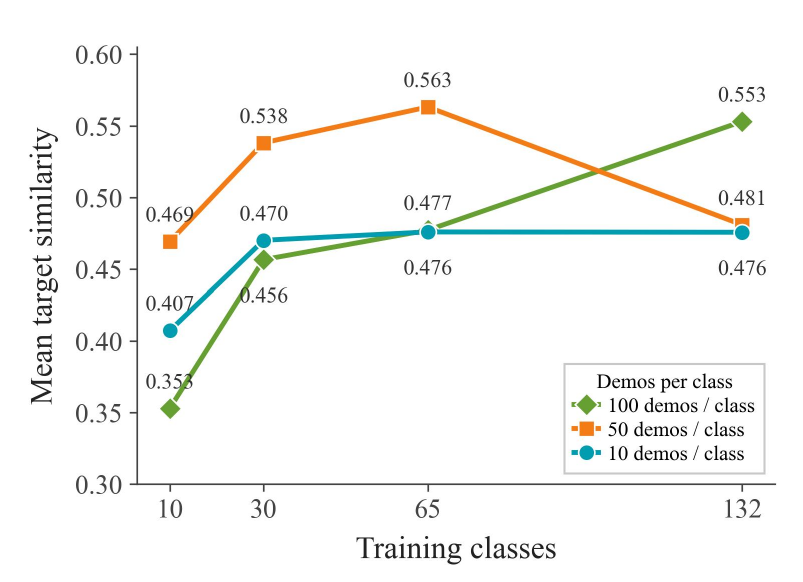
        }}
        \vspace{-15pt}
        \caption{$\pi_{0}$-FAST}
        \label{fig:pi0_fast_scale}
    \end{subfigure}

    \caption{Scaling behavior of $\pi_{0}$, $\pi_{0.5}$, and $\pi_{0}$-FAST across training data scales. The x-axis denotes the number of training classes, the y-axis shows the mean target similarity, and colors indicate the number of demonstrations per class. Flow-matching models improve more consistently with scale, whereas $\pi_{0}$-FAST shows more variable trends.}
    \label{fig:scaling}
    \vspace{-10pt}
\end{figure}

We evaluate the $\pi$-family VLA models across different task scales, varying the number of
object classes from $C=10$ to $C=132$ and the number of trajectories per class
from $T=10$ to $T=100$, as shown in Figure~\ref{fig:scaling}. Drawing performance
is evaluated using the DINOv3-based target similarity metric introduced in
Sec.~\ref{sec:human_aligned_eval}. Overall, increasing either the number of training classes or the
number of trajectories per class generally improves drawing performance,
particularly for $\pi_{0}$ and $\pi_{0.5}$. However, the scaling behavior is not
uniform across models: $\pi_{0}$-FAST shows larger fluctuations and does not
consistently benefit from increasing data scale. These results suggest that
BURGER provides a controlled setting for studying how different VLA formulations
respond to increasing task diversity and demonstration scale. 
\subsection{Language-conditioned Generalization}

\begin{figure*}[t]
    \centering
    \includegraphics[width=\textwidth]{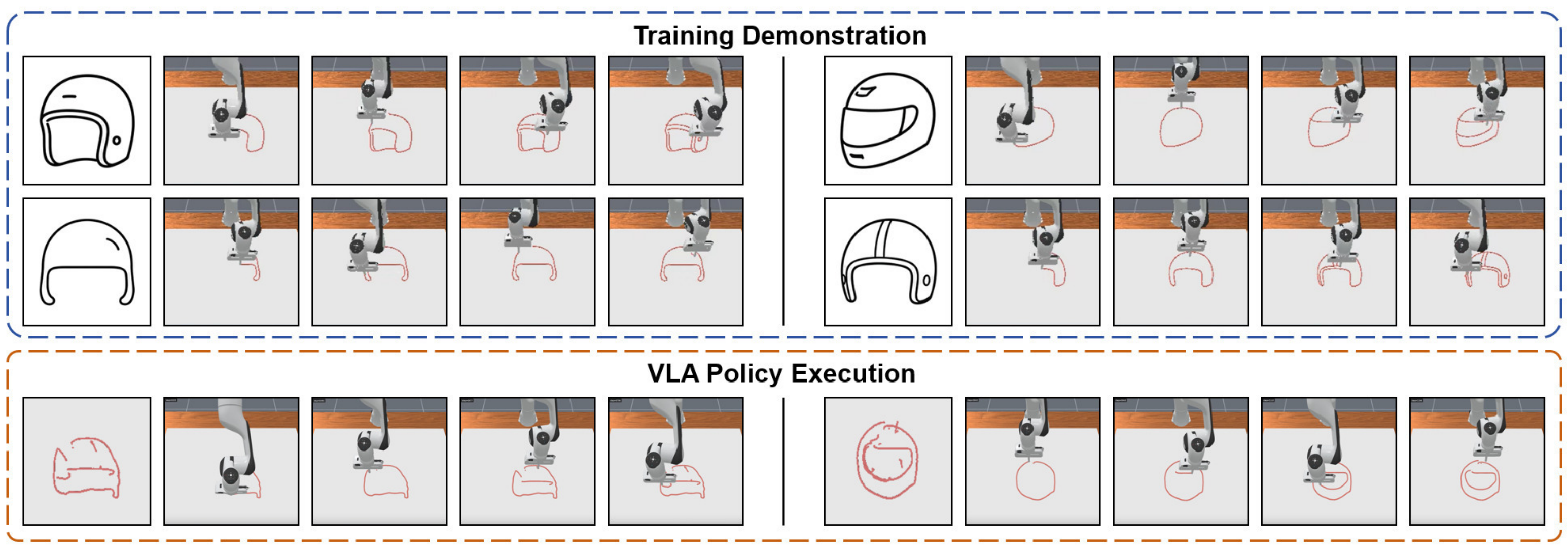}
    \caption{Learning multimodal drawing behaviors in \myenv. \textbf{Top:}
    different drawing patterns of ``helmet'' within the dataset, \textbf{Bottom:}
    the generated images by $\pi_{0.5}$ unseen in the dataset.}
    \label{fig:multi_modal}

    \vspace{-10pt}
\end{figure*}
We diagnose language grounding and semantic generalization across the
$\pi$-family models along two dimensions. First, we evaluate language-conditioned
instruction following by varying the object names in the prompts. Starting from
the same blank canvas, different language instructions lead the policies to
generate different objects corresponding to the requested categories. Across
the in-distribution tasks, we observe no evident confusion between objects. In contrast, when the instruction specifies an out-of-distribution object category, the models fail to generate the
corresponding object. This suggests that the learned language
grounding is effective for the trained semantic categories but does not directly
generalize to unseen object concepts.

Second, we evaluate spatial understanding using compositional drawing tasks with
prompts of the form ``draw A on the left of B'', ``draw A under B'', and
``draw A over B''. Our goal is to study whether VLA models can generalize simple
spatial relations beyond memorized object compositions. For this evaluation,
policies are trained on these compositional tasks. As shown in the right-most
column in Figure~\ref{fig:l}, VLA models can generate valid left-right and
over-under object compositions even for object pairs that are not observed during
training. For example, a policy can draw a cup to the left of an apple despite
never being trained on this specific composition. We also observe that the
policies correctly distinguish different spatial instructions. These results
suggest that the policies can generalize beyond memorized object compositions and
apply learned spatial relations to unseen object pairs. 
\subsection{VLA Models Capture Multimodal Drawing Behaviors}
\label{sec:multimodal_behaviors}
Sketch generation naturally admits multimodal trajectories for the same language
instruction. As quantified in Sec.~\ref{sec:dataset} and visualized in
Figure~\ref{fig:data_comparison}, \myenv~exhibits substantially stronger
within-instruction multimodality than LIBERO. We therefore use it to evaluate
whether VLA policies can learn and preserve such multimodality after training.

We first evaluate this quantitatively by applying the same trajectory representation
and clustering procedure to $\pi_{0}$ rollouts. Across the eight representative
tasks shown in Figure~\ref{fig:data_comparison}, the training demonstrations
exhibit $4.0$--$7.7$ effective trajectory modes per instruction,
while the policy rollouts exhibit $2.1$--$6.7$ modes.
These results show that $\pi_{0}$ can preserve multimodality in its
rollout distribution after training on \myenv.

We further examine this behavior qualitatively in Figure~\ref{fig:multi_modal}.
The first two rows show distinct demonstrations of the same object class,
\emph{helmet}, with variations in geometry, stroke ordering, and execution path.
After training on these demonstrations, $\pi_{0.5}$ produces multiple valid
helmet drawings under the same instruction. Together, the trajectory-level and
visual results show that \myenv~supports learning a distribution of valid drawing
behaviors rather than a single deterministic route.
\subsection{\myenv~Reveals Fine-Grained VLA Failure Modes}
\begin{figure}[ht]
    \centering
    \includegraphics[width=\columnwidth]{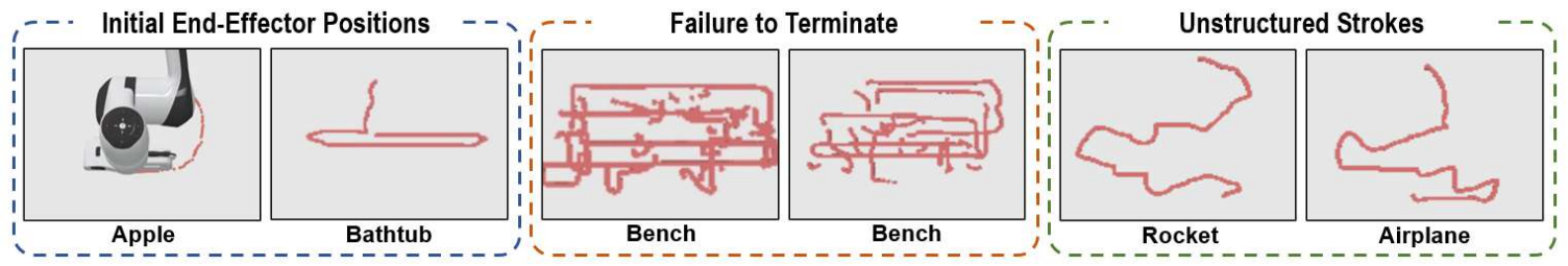}
    \caption{Representative failure modes in BURGER. Left: OpenVLA and $\pi_{0}$-FAST exhibit sensitivity to the initial end-effector position. Middle: $\pi_{0}$ and $\pi_{0.5}$ show failure to terminate. Right: $\pi$-family models produce generic, unstructured strokes on unseen-object instructions.}
    \label{fig:failure_cases_overall}

    \vspace{-10pt}
\end{figure}

Different VLA models exhibit distinct failure patterns, which \myenv~can expose through process-level analysis of generated trajectories and drawings. Figure~\ref{fig:failure_cases_overall} shows some general failure cases.

\textbf{Single-Object Tasks.}
OpenVLA and $\boldsymbol{\pi}_{0}$-FAST are sensitive to the initial end-effector position: OpenVLA may become trapped in a local region, while $\boldsymbol{\pi}_{0}$-FAST can drift away from the intended drawing area. In contrast, flow-matching policies sometimes fail to terminate after completing the object, leading to overdraw. For unseen-object instructions, the policies tend to produce generic, unstructured strokes.

\textbf{Compositional Tasks.}
Object quality may degrade under compositional instructions even when the spatial relation is correct. This degradation becomes more evident when the instruction space expands from horizontal to both horizontal and vertical relations, suggesting that relational grounding and object-level geometry become harder to preserve jointly.

\begin{figure}[ht]
    \centering
    \includegraphics[width=\columnwidth]{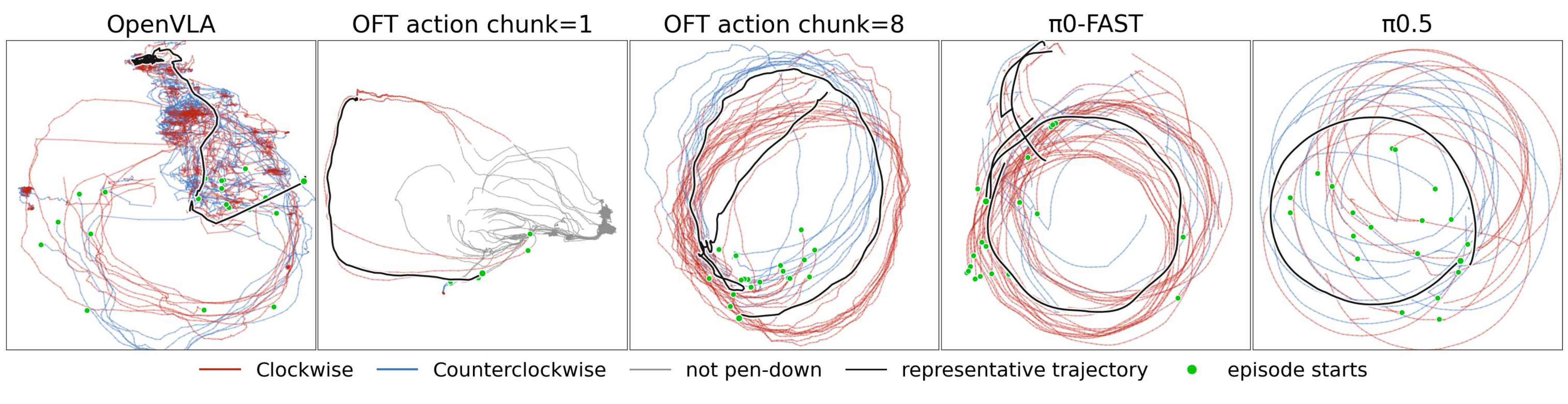}
    \caption{Controlled circle-drawing results across representative VLA models, revealing distinct failure patterns in contact, trajectory progress, and multimodal behavior.}
    \label{fig:circle_task}
    \vspace{-10pt}
\end{figure}

To further diagnose these differences, we construct a controlled circle-drawing case study with 100 demonstrations sharing the same shape and radius, while varying the circle center, starting point, and drawing direction. Table~\ref{tab:circle_diag} summarizes contact, clockwise/counterclockwise (CW/CCW) direction, starting-point diversity, and angular sweep. As shown in Figure~\ref{fig:circle_task}, OpenVLA often maintains contact but stagnates in small local circles, producing excessive angular motion without stable progress along the target trajectory. OpenVLA-OFT with chunk length 1 frequently fails to establish stable contact, while increasing the chunk length to 8 substantially improves contact and trajectory coherence, indicating a clear effect of action chunking on long-horizon execution.

However, coherent execution does not guarantee faithful coverage of the demonstrated behavior distribution. OpenVLA-OFT and $\pi_{0}$-FAST tend to concentrate on a subset of starting points or drawing directions, while the flow-matching policies better preserve trajectory diversity. In particular, $\pi_{0}$-FAST strongly favors one drawing direction despite maintaining relatively accurate circle geometry. Both OpenVLA-OFT and $\pi_{0}$-FAST also exhibit overdraw, continuing beyond the intended single revolution; similar patterns recur across different rollout seeds in Fig.~\ref{fig:circle_task}, suggesting a small set of stereotyped failure behaviors. Overall, these results indicate that action chunking mainly affects execution coherence, while the action-generation formulation also shapes geometric fidelity and multimodal mode coverage. This case study illustrates how the \myenv~pipeline enables controlled diagnosis of model-specific failure patterns.

\begin{table}[ht]
    \centering
    \caption{
    Quantitative results on the controlled circle-drawing task. CW denotes the proportion of clockwise trajectories. Higher start entropy indicates more uniform starting-point coverage, while sweep values closer to $360^\circ$ indicate better completion of a single revolution.}
    \label{tab:circle_diag}
    \small
    \setlength{\tabcolsep}{2.7pt}
    \renewcommand{\arraystretch}{1.10}
    \begin{tabular}{lccccc}
        \toprule
        \textbf{Source}
        & $\mathbf{N}$
        & \textbf{Contact}
        & \textbf{CW(\%)}
        & \textbf{Start Entropy} 
        & \textbf{Sweep($^\circ$)} \\
        \midrule
        Dataset
        & 100 & 100 & 56.0 & 0.97 & 359.8 \\
        OpenVLA
        & 20 & 20 & 40.0 & 0.82 & 874.6 \\
        OFT $H=1$
        & 20 & 6 & 100.0 & 0.35 & 129.3 \\
        OFT $H=8$
        & 20 & 20 & 55.0 & 0.58 & 586.1 \\
        $\pi_{0}$-FAST
        & 20 & 20 & 95.0 & 0.63 & 563.2 \\
        $\pi_{0.5}$
        & 20 & 20 & 55.0 & 0.84 & 355.2 \\
        \bottomrule
    \vspace{-15pt}
    \end{tabular}
\end{table}
\subsection{Real Robot Results}
\label{app:interactive}
\begin{figure*}[ht]
   \centering
   \includegraphics[width=\textwidth]{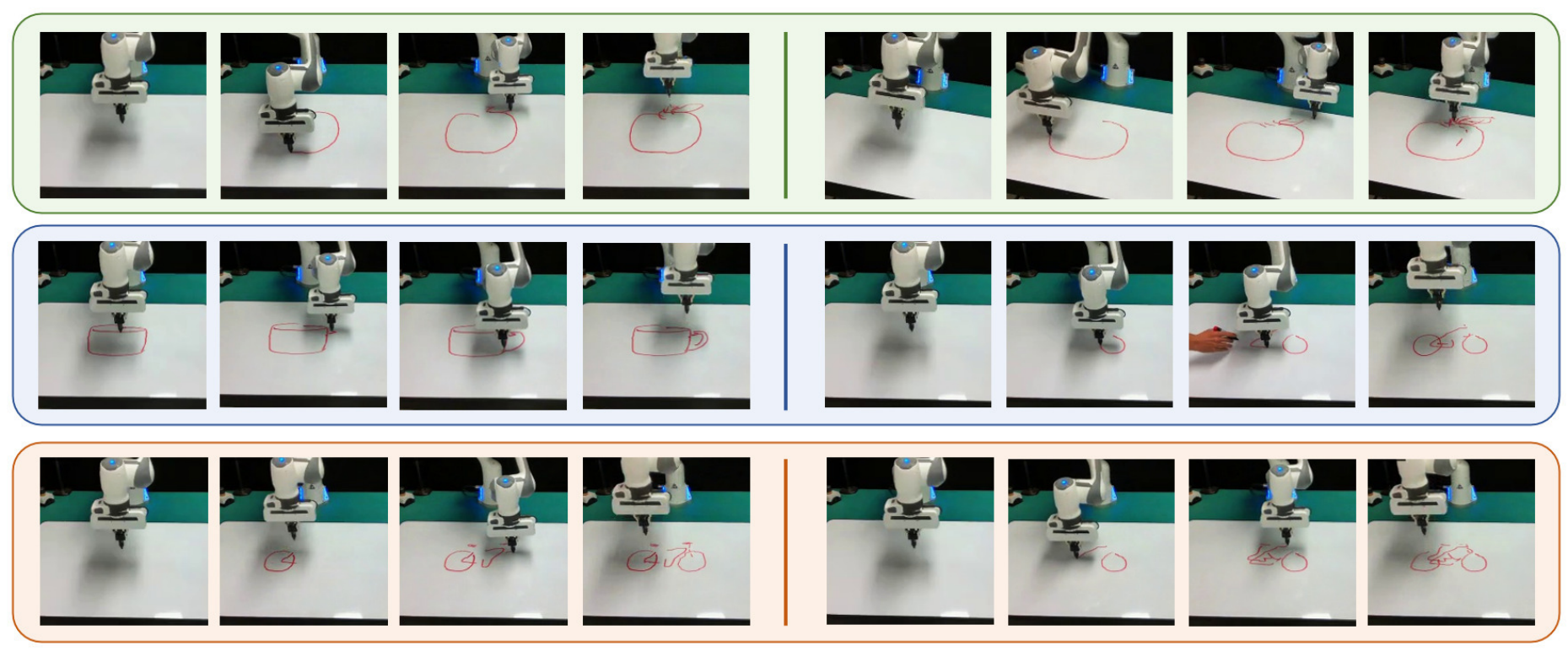}    
   \caption{Real-robot drawing results. The first row shows robustness to camera-angle changes; the second shows visual feedback through object completion and human intervention; the third shows multimodal real-robot behaviors.}
\vspace{-10pt}
\label{fig:real_robot_video} 
\end{figure*}
Using the real-robot demonstrations introduced in Sec.~\ref{sec:dataset}, we fine-tune
$\pi_{0.5}$ directly on data collected with the Franka Panda platform. As shown
in Figure~\ref{fig:real_robot_video}, the resulting policy can reliably execute
real-world drawing tasks, demonstrating that the BURGER data-collection pipeline
can directly support real-world policy learning. The first row further shows that
the policy remains effective under changes in camera viewpoint, indicating moderate
robustness to visual observation variations.

We further examine closed-loop visual feedback through completion and interaction
tasks. Given a partially completed cup, the policy can add the missing handle. In
the bicycle example, after a human draws the rear wheel, the robot observes the
updated canvas, skips the completed component, and continues with the remaining
structure. The third row additionally shows that the policy can produce different
valid trajectories for the same drawing task, preserving multimodal behavior on
the real robot. Together, these results demonstrate visual closed-loop interaction
and multimodal real-world execution.

\section{Conclusion and Limitations}
We present \myenv, a language-conditioned robotic drawing benchmark that supports scalable generation of diverse robot trajectories and fine-grained evaluation of semantic grounding, spatial reasoning, multimodal behavior modeling, and action generation.
Our experiments reveal how representative VLA models respond to increasing data scale and task diversity, while exposing distinct failure patterns beyond binary task success.
Real-robot experiments further demonstrate drawing with visual feedback, including responses to human-added strokes.

Although \myenv~provides a scalable and diagnostic testbed for evaluating language-conditioned robotic drawing, several limitations remain in our benchmark.
First, our setup abstracts away many physical complexities of painting. 
Unlike robotic painting systems that model brush dynamics~\cite{wang2026impasto,schaldenbrand2022frida}, \myenv~focuses on sketch-level stroke generation with a drawing tool. 
This simplification makes the benchmark scalable, but limits its coverage of contact-rich artistic media such as brush painting.
Second, the dataset may inherit biases from the automatic generation pipeline. 
Our stroke-order heuristics may introduce drawing styles that differ from human demonstrations. 
As a result, the dataset may not fully represent natural human drawing behaviors.

\clearpage
\acknowledgments{This work is supported by Tsinghua University Initiative Scientific Research Program (Student Academic Research Advancement Program, 20267020002) and Tsinghua University Dushi Program.}

\bibliography{main}  % .bib

@inproceedings{o2024open,
  title={Open x-embodiment: Robotic learning datasets and rt-x models: Open x-embodiment collaboration 0},
  author={O’Neill, Abby and Rehman, Abdul and Maddukuri, Abhiram and Gupta, Abhishek and Padalkar, Abhishek and Lee, Abraham and Pooley, Acorn and Gupta, Agrim and Mandlekar, Ajay and Jain, Ajinkya and others},
  booktitle={2024 IEEE International Conference on Robotics and Automation (ICRA)},
  pages={6892--6903},
  year={2024},
  organization={IEEE}
}

@article{khazatsky2024droid,
  title={Droid: A large-scale in-the-wild robot manipulation dataset},
  author={Khazatsky, Alexander and Pertsch, Karl and Nair, Suraj and Balakrishna, Ashwin and Dasari, Sudeep and Karamcheti, Siddharth and Nasiriany, Soroush and Srirama, Mohan Kumar and Chen, Lawrence Yunliang and Ellis, Kirsty and others},
  journal={arXiv preprint arXiv:2403.12945},
  year={2024}
}

@article{xing2025shortcut,
  title={Shortcut learning in generalist robot policies: The role of dataset diversity and fragmentation},
  author={Xing, Youguang and Luo, Xu and Xie, Junlin and Gao, Lianli and Shen, Hengtao and Song, Jingkuan},
  journal={arXiv preprint arXiv:2508.06426},
  year={2025}
}

@inproceedings{zitkovich2023rt,
  title={Rt-2: Vision-language-action models transfer web knowledge to robotic control},
  author={Zitkovich, Brianna and Yu, Tianhe and Xu, Sichun and Xu, Peng and Xiao, Ted and Xia, Fei and Wu, Jialin and Wohlhart, Paul and Welker, Stefan and Wahid, Ayzaan and others},
  booktitle={Conference on Robot Learning},
  pages={2165--2183},
  year={2023},
  organization={PMLR}
}

@article{kim2024openvla,
  title={Openvla: An open-source vision-language-action model},
  author={Kim, Moo Jin and Pertsch, Karl and Karamcheti, Siddharth and Xiao, Ted and Balakrishna, Ashwin and Nair, Suraj and Rafailov, Rafael and Foster, Ethan and Lam, Grace and Sanketi, Pannag and others},
  journal={arXiv preprint arXiv:2406.09246},
  year={2024}
}

@misc{black2026pi0visionlanguageactionflowmodel,
      title={$\pi_0$: A Vision-Language-Action Flow Model for General Robot Control}, 
      author={Kevin Black and Noah Brown and Danny Driess and Adnan Esmail and Michael Equi and Chelsea Finn and Niccolo Fusai and Lachy Groom and Karol Hausman and Brian Ichter and Szymon Jakubczak and Tim Jones and Liyiming Ke and Sergey Levine and Adrian Li-Bell and Mohith Mothukuri and Suraj Nair and Karl Pertsch and Lucy Xiaoyang Shi and James Tanner and Quan Vuong and Anna Walling and Haohuan Wang and Ury Zhilinsky},
      year={2026},
      eprint={2410.24164},
      archivePrefix={arXiv},
      primaryClass={cs.LG},
      url={https://arxiv.org/abs/2410.24164}, 
}

@inproceedings{zhang2025vlabench,
  title={Vlabench: A large-scale benchmark for language-conditioned robotics manipulation with long-horizon reasoning tasks},
  author={Zhang, Shiduo and Xu, Zhe and Liu, Peiju and Yu, Xiaopeng and Li, Yuan and Gao, Qinghui and Fei, Zhaoye and Yin, Zhangyue and Wu, Zuxuan and Jiang, Yu-Gang and others},
  booktitle={Proceedings of the IEEE/CVF International Conference on Computer Vision},
  pages={11142--11152},
  year={2025}
}

@article{kim2025fine,
  title={Fine-tuning vision-language-action models: Optimizing speed and success},
  author={Kim, Moo Jin and Finn, Chelsea and Liang, Percy},
  journal={arXiv preprint arXiv:2502.19645},
  year={2025}
}

@article{wang2025roboeval,
  title={Roboeval: Where robotic manipulation meets structured and scalable evaluation},
  author={Wang, Yi Ru and Ung, Carter and Tan, Christopher and Tannert, Grant and Duan, Jiafei and Li, Josephine and Le, Anh and Oswal, Rishabh and Grotz, Markus and Pumacay, Wilbert and others},
  journal={arXiv preprint arXiv:2507.00435},
  year={2025}
}

@article{mees2022calvin,
  title={Calvin: A benchmark for language-conditioned policy learning for long-horizon robot manipulation tasks},
  author={Mees, Oier and Hermann, Lukas and Rosete-Beas, Erick and Burgard, Wolfram},
  journal={IEEE Robotics and Automation Letters},
  volume={7},
  number={3},
  pages={7327--7334},
  year={2022},
  publisher={IEEE}
}

@article{liu2023libero,
  title={Libero: Benchmarking knowledge transfer for lifelong robot learning},
  author={Liu, Bo and Zhu, Yifeng and Gao, Chongkai and Feng, Yihao and Liu, Qiang and Zhu, Yuke and Stone, Peter},
  journal={Advances in Neural Information Processing Systems},
  volume={36},
  pages={44776--44791},
  year={2023}
}

@article{pertsch2025fast,
  title={Fast: Efficient action tokenization for vision-language-action models},
  author={Pertsch, Karl and Stachowicz, Kyle and Ichter, Brian and Driess, Danny and Nair, Suraj and Vuong, Quan and Mees, Oier and Finn, Chelsea and Levine, Sergey},
  journal={arXiv preprint arXiv:2501.09747},
  year={2025}
}

@article{taomaniskill3,
  title={ManiSkill3: GPU Parallelized Robotics Simulation and Rendering for Generalizable Embodied AI},
  author={Stone Tao and Fanbo Xiang and Arth Shukla and Yuzhe Qin and Xander Hinrichsen and Xiaodi Yuan and Chen Bao and Xinsong Lin and Yulin Liu and Tse-kai Chan and Yuan Gao and Xuanlin Li and Tongzhou Mu and Nan Xiao and Arnav Gurha and Viswesh Nagaswamy Rajesh and Yong Woo Choi and Yen-Ru Chen and Zhiao Huang and Roberto Calandra and Rui Chen and Shan Luo and Hao Su},
  journal = {Robotics: Science and Systems},
  year={2025},
}

@article{zhang2023text,
  title={Text-to-image diffusion models in generative ai: A survey},
  author={Zhang, Chenshuang and Zhang, Chaoning and Zhang, Mengchun and Kweon, In So and Kim, Junmo},
  journal={arXiv preprint arXiv:2303.07909},
  year={2023}
}

@misc{openai_gpt_image_2,
  author       = {{OpenAI}},
  title        = {GPT-Image-2},
  year         = {2026},
  howpublished = {\url{https://platform.openai.com/docs/models\#gpt-image-2}},
}

@article{simeoni2025dinov3,
  title={Dinov3},
  author={Sim{\'e}oni, Oriane and Vo, Huy V and Seitzer, Maximilian and Baldassarre, Federico and Oquab, Maxime and Jose, Cijo and Khalidov, Vasil and Szafraniec, Marc and Yi, Seungeun and Ramamonjisoa, Micha{\"e}l and others},
  journal={arXiv preprint arXiv:2508.10104},
  year={2025}
}

@inproceedings{hessel2021clipscore,
  title={Clipscore: A reference-free evaluation metric for image captioning},
  author={Hessel, Jack and Holtzman, Ari and Forbes, Maxwell and Le Bras, Ronan and Choi, Yejin},
  booktitle={Proceedings of the 2021 conference on empirical methods in natural language processing},
  pages={7514--7528},
  year={2021}
}

@article{ahmadi2023robustness,
  title={An Examination of the Robustness of Reference-Free Image Captioning Evaluation Metrics},
  author={Ahmadi, Saba and Agrawal, Aishwarya},
  journal={arXiv preprint arXiv:2305.14998},
  year={2023}
}

@article{wang2025vqvla,
  title = {{VQ-VLA}: Improving Vision-Language-Action Models via Scaling
           Vector-Quantized Action Tokenizers},
  author = {Wang, Yating and Zhu, Haoyi and Liu, Mingyu
            and Yang, Jiange and Fang, Hao-Shu and He, Tong},
  journal = {arXiv preprint arXiv:2507.01016},
  year = {2025},
  url = {https://arxiv.org/abs/2507.01016}
}

@article{liu2025faster,
  title = {{FASTer}: Toward Efficient Autoregressive Vision-Language-Action
           Modeling via Neural Action Tokenization},
  author = {Liu, Yicheng and Zhang, Shiduo and Dong, Zibin
            and Ye, Baijun and Yuan, Tianyuan and Yu, Xiaopeng
            and Yin, Linqi and Lu, Chenhao and Shi, Junhao
            and Yu, Luca Jiang-Tao and Zheng, Liangtao
            and Jiang, Tao and Gong, Jingjing and Qiu, Xipeng
            and Zhao, Hang},
  journal = {arXiv preprint arXiv:2512.04952},
  year = {2025},
  url = {https://arxiv.org/abs/2512.04952}
}

@article{song2025pdvla,
  title = {{PD-VLA}: Accelerating Vision-Language-Action Model Integrated
           with Action Chunking via Parallel Decoding},
  author = {Song, Wenxuan and Chen, Jiayi and Ding, Pengxiang
            and Zhao, Han and Zhao, Wei and Zhong, Zhide
            and Ge, Zongyuan and Li, Zhijun and Wang, Donglin
            and Ma, Jun and Wang, Lujia and Li, Haoang},
  journal = {arXiv preprint arXiv:2503.02310},
  year = {2025},
  url = {https://arxiv.org/abs/2503.02310}
}

@InProceedings{black2025pi05,
  title = 	 {$\pi_{0.5}$: a Vision-Language-Action Model with Open-World Generalization},
  author =       {Black, Kevin and Brown, Noah and Darpinian, James and Dhabalia, Karan and Driess, Danny and Esmail, Adnan and Equi, Michael Robert and Finn, Chelsea and Fusai, Niccolo and Galliker, Manuel Y. and Ghosh, Dibya and Groom, Lachy and Hausman, Karol and ichter, brian and Jakubczak, Szymon and Jones, Tim and Ke, Liyiming and LeBlanc, Devin and Levine, Sergey and Li-Bell, Adrian and Mothukuri, Mohith and Nair, Suraj and Pertsch, Karl and Ren, Allen Z. and Shi, Lucy Xiaoyang and Smith, Laura and Springenberg, Jost Tobias and Stachowicz, Kyle and Tanner, James and Vuong, Quan and Walke, Homer and Walling, Anna and Wang, Haohuan and Yu, Lili and Zhilinsky, Ury},
  booktitle = 	 {Proceedings of The 9th Conference on Robot Learning},
  pages = 	 {17--40},
  year = 	 {2025},
  editor = 	 {Lim, Joseph and Song, Shuran and Park, Hae-Won},
  volume = 	 {305},
  series = 	 {Proceedings of Machine Learning Research},
  month = 	 {27--30 Sep},
  publisher =    {PMLR},
  url = 	 {https://proceedings.mlr.press/v305/black25a.html}
}

@inproceedings{
liu2025rdt,
title={{RDT}-1B: a Diffusion Foundation Model for Bimanual Manipulation},
author={Songming Liu and Lingxuan Wu and Bangguo Li and Hengkai Tan and Huayu Chen and Zhengyi Wang and Ke Xu and Hang Su and Jun Zhu},
booktitle={The Thirteenth International Conference on Learning Representations},
year={2025},
url={https://openreview.net/forum?id=yAzN4tz7oI}
}

@article{bjorck2025gr00tn1,
  title={Gr00t n1: An open foundation model for generalist humanoid robots},
  author={Bjorck, Johan and Casta{\~n}eda, Fernando and Cherniadev, Nikita and Da, Xingye and Ding, Runyu and Fan, Linxi and Fang, Yu and Fox, Dieter and Hu, Fengyuan and Huang, Spencer and others},
  journal={arXiv preprint arXiv:2503.14734},
  year={2025}
}

@InProceedings{wen2025diffusionvla,
  title = 	 {{D}iffusion{VLA}: Scaling Robot Foundation Models via Unified Diffusion and Autoregression},
  author =       {Wen, Junjie and Zhu, Yichen and Zhu, Minjie and Tang, Zhibin and Li, Jinming and Zhou, Zhongyi and Liu, Xiaoyu and Shen, Chaomin and Peng, Yaxin and Feng, Feifei},
  booktitle = 	 {Proceedings of the 42nd International Conference on Machine Learning},
  pages = 	 {66558--66574},
  year = 	 {2025},
  editor = 	 {Singh, Aarti and Fazel, Maryam and Hsu, Daniel and Lacoste-Julien, Simon and Berkenkamp, Felix and Maharaj, Tegan and Wagstaff, Kiri and Zhu, Jerry},
  volume = 	 {267},
  series = 	 {Proceedings of Machine Learning Research},
  month = 	 {13--19 Jul},
  publisher =    {PMLR},
  url = 	 {https://proceedings.mlr.press/v267/wen25g.html}
}

@inproceedings{
lipman2023flowmatching,
title={Flow Matching for Generative Modeling},
author={Yaron Lipman and Ricky T. Q. Chen and Heli Ben-Hamu and Maximilian Nickel and Matthew Le},
booktitle={The Eleventh International Conference on Learning Representations },
year={2023},
url={https://openreview.net/forum?id=PqvMRDCJT9t}
}

@article{wang2026impasto,
  title={IMPASTO: Integrating Model-Based Planning with Learned Dynamics Models for Robotic Oil Painting Reproduction},
  author={Wang, Yingke and Li, Hao and Zhu, Yifeng and Yu, Hong-Xing and Goldberg, Ken and Fei-Fei, Li and Wu, Jiajun and Li, Yunzhu and Zhang, Ruohan},
  journal={arXiv preprint arXiv:2603.29315},
  year={2026}
}

@article{wang2025vlatest,
  title={Vlatest: Testing and evaluating vision-language-action models for robotic manipulation},
  author={Wang, Zhijie and Zhou, Zhehua and Song, Jiayang and Huang, Yuheng and Shu, Zhan and Ma, Lei},
  journal={Proceedings of the ACM on Software Engineering},
  volume={2},
  number={FSE},
  pages={1615--1638},
  year={2025},
  publisher={ACM New York, NY, USA}
}

@article{dai2026robomme,
  title={Robomme: Benchmarking and understanding memory for robotic generalist policies},
  author={Dai, Yinpei and Fu, Hongze and Lee, Jayjun and Liu, Yuejiang and Zhang, Haoran and Yang, Jianing and Finn, Chelsea and Fazeli, Nima and Chai, Joyce},
  journal={arXiv preprint arXiv:2603.04639},
  year={2026}
}

@article{chen2026rmbench,
  title={Rmbench: Memory-dependent robotic manipulation benchmark with insights into policy design},
  author={Chen, Tianxing and Wang, Yuran and Li, Mingleyang and Qin, Yan and Shi, Hao and Li, Zixuan and Hu, Yifan and Zhang, Yingsheng and Wang, Kaixuan and Chen, Yue and others},
  journal={arXiv preprint arXiv:2603.01229},
  year={2026}
}

@inproceedings{lee2022scratch,
  title={From scratch to sketch: Deep decoupled hierarchical reinforcement learning for robotic sketching agent},
  author={Lee, Ganghun and Kim, Minji and Lee, Minsu and Zhang, Byoung-Tak},
  booktitle={2022 International Conference on Robotics and Automation (ICRA)},
  pages={5553--5559},
  year={2022},
  organization={IEEE}
}

@article{schaldenbrand2022frida,
  title={Frida: A collaborative robot painter with a differentiable, real2sim2real planning environment},
  author={Schaldenbrand, Peter and McCann, James and Oh, Jean},
  journal={arXiv preprint arXiv:2210.00664},
  year={2022}
}

@inproceedings{
lin2025datascaling,
title={Data Scaling Laws in Imitation Learning for Robotic Manipulation},
author={Fanqi Lin and Yingdong Hu and Pingyue Sheng and Chuan Wen and Jiacheng You and Yang Gao},
booktitle={The Thirteenth International Conference on Learning Representations},
year={2025},
url={https://openreview.net/forum?id=pISLZG7ktL}
}

@inproceedings{
saxena2025whatmatters,
title={What Matters in Learning from Large-Scale Datasets for Robot Manipulation},
author={Vaibhav Saxena and Matthew Bronars and Nadun Ranawaka Arachchige and Kuancheng Wang and Woo Chul Shin and Soroush Nasiriany and Ajay Mandlekar and Danfei Xu},
booktitle={The Thirteenth International Conference on Learning Representations},
year={2025},
url={https://openreview.net/forum?id=LqhorpRLIm}
}

@article{david2017neural,
  author    = {David Ha and
               Douglas Eck},
  title     = {A Neural Representation of Sketch Drawings},
  journal   = {CoRR},
  volume    = {abs/1704.03477},
  year      = {2017},
  url       = {http://arxiv.org/abs/1704.03477},
  archivePrefix = {arXiv},
  eprint    = {1704.03477},
  bibsource = {dblp computer science bibliography, https://dblp.org}
}

@misc{google2026gemini31pro,
  title        = {Gemini 3.1 Pro Model Card},
  author       = {{Google DeepMind}},
  year         = {2026},
  howpublished = {\url{https://deepmind.google/models/model-cards/gemini-3-1-pro/}},
  note         = {Accessed: 2026-06-04}
}

@InProceedings{vinker2025sketchagent,
            author    = {Vinker, Yael and Shaham, Tamar Rott and Zheng, Kristine and Zhao, Alex and E Fan, Judith and Torralba, Antonio},
            title     = {SketchAgent: Language-Driven Sequential Sketch Generation},
            booktitle = {Proceedings of the Computer Vision and Pattern Recognition Conference (CVPR)},
            month     = {June},
            year      = {2025},
            pages     = {23355-23368}
        }

@article{du2026sketch,
  title   = {Teaching an Agent to Sketch One Part at a Time},
  author  = {Du, Xiaodan and Xu, Ruize and Yunis, David and Vinker, Yael and Shakhnarovich, Greg},
  journal = {arXiv preprint arXiv:2603.19500},
  year    = {2026}
}

@inproceedings{sdxl,
 author = {Podell, Dustin and English, Zion and Lacey, Kyle and Blattmann, Andreas and Dockhorn, Tim and M\"{u}ller, Jonas and Penna, Joe and Rombach, Robin},
 booktitle = {International Conference on Learning Representations},
 editor = {B. Kim and Y. Yue and S. Chaudhuri and K. Fragkiadaki and M. Khan and Y. Sun},
 pages = {1862--1874},
 title = {SDXL: Improving Latent Diffusion Models for High-Resolution Image Synthesis},
 url = {https://proceedings.iclr.cc/paper_files/paper/2024/file/081b08068e4733ae3e7ad019fe8d172f-Paper-Conference.pdf},
 volume = {2024},
 year = {2024}
}

@inproceedings{swiftsketch,
author = {Arar, Ellie and Frenkel, Yarden and Cohen-Or, Daniel and Shamir, Ariel and Vinker, Yael},
title = {SwiftSketch: A Diffusion Model for Image-to-Vector Sketch Generation},
year = {2025},
isbn = {9798400715402},
publisher = {Association for Computing Machinery},
address = {New York, NY, USA},
url = {https://doi.org/10.1145/3721238.3730612},
doi = {10.1145/3721238.3730612},
booktitle = {Proceedings of the Special Interest Group on Computer Graphics and Interactive Techniques Conference Conference Papers},
articleno = {82},
numpages = {12},
location = {
},
series = {SIGGRAPH Conference Papers '25}
}

@misc{cadene2024lerobot,
    author = {Cadene, Remi and Alibert, Simon and Soare, Alexander and Gallouedec, Quentin and Zouitine, Adil and Palma, Steven and Kooijmans, Pepijn and Aractingi, Michel and Shukor, Mustafa and Aubakirova, Dana and Russi, Martino and Capuano, Francesco and Pascal, Caroline and Choghari, Jade and Meftah, Khalil and Ellerbach, Maxime and Moss, Jess and Wolf, Thomas},
    title = {LeRobot: State-of-the-art Machine Learning for Real-World Robotics in Pytorch},
    howpublished = "\url{https://github.com/huggingface/lerobot}",
    year = {2024}
}
\newpage
\appendix
\section*{Appendix}

\section{Dataset Details}

\subsection{Prompting and Sketch Generation}
\label{app:png_generation}

\textbf{PNG Generation.} We use GPT-Image 2~\cite{openai_gpt_image_2} to generate
sketch images from language prompts. To improve generation efficiency, each query
produces a $4 \times 4$ grid of subfigures, which is later split into individual
PNG sketches. The image-generation prompt is shown in Figure \ref{fig:png_generation_prompt}.
We substitute the object name into a fixed template, while optionally varying the
diversity instruction to obtain multiple minimal line-art candidates. The
generated grid image is subsequently split into individual PNG sketches.

\begin{figure*}[!b]
    \centering
    \fbox{
    \begin{minipage}{0.96\linewidth}
        \vspace{0.4em}
        \textbf{Prompt sent to the image generator:}

        \vspace{0.3em}
        \footnotesize Create one image containing exactly 16 ultra-simple black
        line-art \texttt{<object>} icons arranged in a clean $4$ by $4$ grid.
        Each cell contains one centered icon with generous white space. All icons
        must be recognizable as the same object category, and their basic outer
        shape should remain mostly consistent. Every icon should still look slightly
        different through small pose, size, angle, or detail changes.

        Important view constraint: regardless of pose or angle, the object's
        front/main recognizable face must be mostly visible. Avoid rear views,
        back views, extreme side views, hidden fronts, heavy occlusion, or
        cropped-off main faces.

        Make diversity come from combining several simple differences, without adding
        line complexity. Use clear size variation across the grid, but keep every
        icon isolated and easy to cut out. For animals, combine pose, slight
        front-facing angle, head direction, tail/ear shape, and body proportion changes.
        For objects and still lifes, combine front-facing orientation,
        silhouette, placement, proportion, and one tiny detail change.

        Use these variation types: small silhouette differences, slightly different
        sizes, small pose or action changes for animals, small orientation or placement
        changes for objects and still lifes, slight viewing angle changes, and one
        tiny contour/detail change.

        Avoid repeating the same design twice. Do not create very different
        object types, complex shapes, dense internal lines, realistic texture,
        horizontal compositions, extreme tall shapes, extreme wide shapes, and extra
        props.

        Hard line-count constraint: each icon should be composed of only 5 or 6
        total simple strokes/lines whenever possible. Use only the outer contour
        plus at most 1 or 2 tiny internal detail lines. Avoid many details, many
        small marks, texture, shading, filling, decorations, dense internal
        lines, or line-heavy patterns. Each icon should be a clean minimal
        outline, easy to cut out, with a compact simple shape. Black lines only,
        white background, no color, no shading, no filled areas except tiny solid
        black dots used only for animal eyes as specified below, no text, no watermark,
        no extra objects. Each icon should fit comfortably inside an imaginary square
        cell with similar margins on all sides.

        Category-specific rules: For animals, always draw the full whole body in
        each cell—never head-only, face crop, or partial-body views; every
        animal must have clearly visible eyes (never omit eyes). For food, never
        show bites or eaten/missing chunks; do not draw plates, bowls, or other dishware;
        if the food has cream, frosting, or similar topping, keep the same topping
        type and overall shape across every icon (only its scale or amount may
        vary slightly). For furniture, clothing, or animals, every icon must stay
        the same specific type within that category (one furniture piece kind, one
        garment kind, or one animal kind—do not mix different types). For
        animals whose eyes would normally be a large solid black patch or filled
        black area, render each eye as a single small black dot instead. For
        simple symbolic or single-form subjects with only one natural type (for example
        diamond, envelope, eye, star, or heart), keep that exact same shape and meaning
        in every cell—do not invent alternate variants— and create variety only through
        slight rotation, tilt, mirroring, or small in-plane skew.

        \vspace{0.4em}
        \textbf{Output:} one PNG image containing a $4 \times 4$ grid of
        minimal sketch icons.
    \end{minipage}
    }
    \caption{Prompt template for PNG sketch generation.}
    \label{fig:png_generation_prompt}
\end{figure*}

\subsection{Data-Generation Algorithms}
\label{app:algorithms}

\textbf{PNG to Skeleton.} In Algorithm \ref{alg:png_to_skeleton}, we binarize
and skeletonize the PNG image to obtain a skeleton figure.

\begin{algorithm*}
    [ht]
    \caption{PNG-to-Skeleton Conversion}
    \label{alg:png_to_skeleton}
    \begin{algorithmic}
        [1] \State \textbf{Input:} PNG folder $\mathcal{D}_{\mathrm{png}}$, output
        skeleton folder $\mathcal{D}_{\mathrm{skel}}$

        \For{each PNG image $I$ in $\mathcal{D}_{\mathrm{png}}$} \State Load $I$
        as a grayscale image.
        \State Binarize the image into foreground strokes and background:
        \Statex \hspace{\algorithmicindent} dark regions as foreground and white
        regions as background.

        \State Detect connected dark contours in the binary image. \For{each detected contour}
        \State Compute its area, perimeter, and circularity. \If{the contour has valid area and is approximately circular}
        \State Compute the contour centroid. \State Suppress each filled dark region
        by drawing a small white circle at its centroid. \EndIf \EndFor

        \State Apply morphological skeletonization to the cleaned binary foreground.
        \State Convert the skeleton to a binary image format and save to
        $\mathcal{D}_{\mathrm{skel}}$. \EndFor \State \textbf{Output:}
        Skeletonized binary images saved to $\mathcal{D}_{\mathrm{skel}}$
    \end{algorithmic}
\end{algorithm*}

\textbf{Skeleton to SVG.} Algorithm \ref{alg:skeleton_to_svg} summarizes the
procedure of skeleton-to-SVG conversion. We first decompose the skeleton into ordered
stroke paths and then convert each path into SVG commands. The resulting SVG
representation preserves the stroke structure and ordering, and is subsequently replayed
in the robot drawing environment to generate observation--action trajectories.

\begin{algorithm*}
    [ht]
    \caption{Skeleton-to-SVG Conversion}
    \label{alg:skeleton_to_svg}
    \begin{algorithmic}
        [1] \State \textbf{Input:} Skeleton folder $\mathcal{D}_{\mathrm{skel}}$,
        output SVG folder $\mathcal{D}_{\mathrm{svg}}$, sampling spacing
        $\Delta$

        \For{each skeleton image $S$ in $\mathcal{D}_{\mathrm{skel}}$} \State
        Load $S$ as a binary skeleton image. \State Initialize an undirected graph
        $G$.

        \For{each foreground pixel $(x,y)$ in $S$} \For{each 8-connected neighboring pixel $(x',y')$}
        \If{$(x',y')$ is also a foreground pixel} \State Add an edge between
        $(x,y)$ and $(x',y')$ in $G$. \EndIf \EndFor \EndFor

        \State Extract disjoint raw stroke paths from $G$ by traversing
        unvisited edges.
        \For{each path $P$} \State Compute its arc length. \State Uniformly
        sample points along $P$ using spacing $\Delta$. \If{$P$ is approximately circular}
        \State Smooth the sampled points using spline interpolation. \EndIf \EndFor

        \State Sort the sampled paths into a human-like drawing order: \Statex \hspace{\algorithmicindent}
        first draw large outline paths based on size and length; \Statex
        \hspace{\algorithmicindent} then draw smaller detail paths from top to
        bottom using local nearest-neighbor ordering.

        \State Initialize an SVG canvas with the same width and height as $S$.

        \For{each ordered path $P=\{(x_{1},y_{1}),\ldots,(x_{n},y_{n})\}$}
        \State Convert $P$ into an SVG path command.
        \EndFor

        \State Save the SVG file to $\mathcal{D}_{\mathrm{svg}}$. \EndFor \State
        \textbf{Output:} SVG files saved to $\mathcal{D}_{\mathrm{svg}}$
    \end{algorithmic}
\end{algorithm*}

\subsection{Dataset Composition and Statistics}
\label{app:dataset_stat}

The \myenv~dataset contains 132 single-object tasks and 150 compositional tasks.
The full list of single-object classes is provided in Table~\ref{tab:single_object_classes},
and the 100 horizontal compositional tasks are provided in Table~\ref{tab:compositional_tasks}.
The remaining 50 compositional tasks use vertical relations such as ``over'' and ``under.''
Each task contains 100 demonstrations generated through the pipeline described
in Section~\ref{app:algorithms}, for a total of 28,200 simulated demonstrations.

\begin{table*}
    [h]
    \centering
    \small
    \setlength{\tabcolsep}{4pt}
    \renewcommand{\arraystretch}{1.12}
    \caption{ Single-object classes in \myenv. }
    \label{tab:single_object_classes}
    \begin{tabular}{p{0.3\linewidth} p{0.6\linewidth}}
        \toprule \textbf{Category} & \textbf{Classes}  \\
        \midrule
Animals (14)               & \texttt{butterfly}, \texttt{cat}, \texttt{cow}, \texttt{dog}, \texttt{duck}, \texttt{elephant}, \texttt{fish}, \texttt{frog}, \texttt{horse}, \texttt{owl}, \texttt{penguin}, \texttt{pig}, \texttt{rabbit}, \texttt{zebra}                                                                                                \\
        Food (17)                 & \texttt{apple}, \texttt{banana}, \texttt{birthday\_cake}, \texttt{blueberry}, \texttt{donut}, \texttt{grapes}, \texttt{hot\_dog}, \texttt{ice\_cream}, \texttt{lollipop}, \texttt{mushroom}, \texttt{onion}, \texttt{pear}, \texttt{pineapple}, \texttt{pizza}, \texttt{popsicle}, \texttt{sandwich}, \texttt{strawberry}  \\
        Kitchenware (10)       & \texttt{bottlecap}, \texttt{bucket}, \texttt{coffee\_cup}, \texttt{cup}, \texttt{mug}, \texttt{teapot}, \texttt{wine\_bottle}, \texttt{fork},\texttt{knife},    \texttt{spoon},                                                                                                                                                                                           \\
        Vehicles and transport (11)          & \texttt{bicycle}, \texttt{bus}, \texttt{car}, \texttt{police\_car}, \texttt{sailboat}, \texttt{school\_bus}, \texttt{skateboard}, \texttt{traffic\_light}, \texttt{train}, \texttt{truck}, \texttt{wheel}                                                                                                                  \\
        Furniture and home objects (17)      & \texttt{bathtub}, \texttt{bed}, \texttt{bench}, \texttt{candle}, \texttt{chair}, \texttt{couch}, \texttt{door}, \texttt{dresser}, \texttt{house}, \texttt{mailbox}, \texttt{picture\_frame}, \texttt{pillow}, \texttt{sink}, \texttt{table}, \texttt{tent}, \texttt{toilet}, \texttt{vase}                                 \\
        Clothing and accessories (18)        & \texttt{backpack}, \texttt{belt}, \texttt{bowtie}, \texttt{crown}, \texttt{eyeglasses}, \texttt{hat}, \texttt{headphones}, \texttt{helmet}, \texttt{jacket}, \texttt{necklace}, \texttt{pants}, \texttt{purse}, \texttt{shoe}, \texttt{sock}, \texttt{suitcase}, \texttt{t\_shirt}, \texttt{umbrella}, \texttt{wristwatch} \\
        Tools and stationery (11) & \texttt{book}, \texttt{crayon}, \texttt{envelope}, \texttt{eraser},  \texttt{hammer}, \texttt{key}, \texttt{paintbrush}, \texttt{paper\_clip}, \texttt{pencil}, \texttt{scissors}, \texttt{toothbrush}                                                                       \\
        Electronics and appliances (11)      & \texttt{alarm\_clock}, \texttt{camera}, \texttt{cell\_phone}, \texttt{computer}, \texttt{electrical\_fan}, \texttt{laptop}, \texttt{light\_bulb}, \texttt{radio}, \texttt{remote\_control}, \texttt{television}, \texttt{toaster}                                                                                          \\
        Nature and sky (11)                  & \texttt{cloud}, \texttt{flower}, \texttt{leaf}, \texttt{moon}, \texttt{mountain}, \texttt{palm\_tree}, \texttt{rainbow}, \texttt{snowman}, \texttt{star}, \texttt{sun}, \texttt{tree}                                                                                                                                      \\
        Body parts and symbols (7)           & \texttt{diamond}, \texttt{ear}, \texttt{foot}, \texttt{hand}, \texttt{nose}, \texttt{smiley\_face}, \texttt{tooth}                                                                                                                                                                                                         \\
        Sports and music (5)                 & \texttt{basketball}, \texttt{baseball}, \texttt{guitar}, \texttt{piano}, \texttt{soccer\_ball}                                                                                                                                                                                                                             \\
        \bottomrule
    \end{tabular}
\end{table*}

\begin{table*}
    [ht]
    \centering
    \scriptsize
    \setlength{\tabcolsep}{4pt}
    \renewcommand{\arraystretch}{1.08}
    \caption{ Compositional tasks in \myenv. Each listed task follows the form \texttt{A\_left\_B},
    corresponding to the language instruction ``draw A to the left of B.'' The table lists
    the 100 horizontal tasks by the left object $A$ and the corresponding right-side objects $B$;
    the dataset also contains 50 tasks with vertical relations. }
    \label{tab:compositional_tasks}
    \begin{tabular}{p{0.18\linewidth} p{0.74\linewidth} c}
        \toprule \textbf{Left object $A$} & \textbf{Right-side objects $B$}                                                                           & \textbf{\#} \\
        \midrule

\texttt{leaf}           & \texttt{lollipop}, \texttt{mailbox}, \texttt{moon}, \texttt{mountain}, \texttt{snowman}                   & 5           \\
        \texttt{light\_bulb}              & \texttt{leaf}, \texttt{picture\_frame}, \texttt{radio}                                                    & 3           \\
        \texttt{lollipop}                 & \texttt{leaf}, \texttt{toaster}                                                                           & 2           \\
        \texttt{mailbox}                  & \texttt{leaf}, \texttt{moon}, \texttt{rainbow}                                                            & 3           \\
        \texttt{moon}                     & \texttt{leaf}, \texttt{mailbox}, \texttt{paintbrush}, \texttt{sailboat}                                   & 4           \\
        \texttt{mountain}                 & \texttt{owl}                                                                                              & 1           \\
        \texttt{mug}                      & \texttt{piano}, \texttt{pineapple}, \texttt{pizza}                                                        & 3           \\
        \texttt{necklace}                 & \texttt{scissors}, \texttt{truck}                                                                         & 2           \\
        \texttt{nose}                     & \texttt{shoe}                                                                                             & 1           \\
        \texttt{onion}                    & \texttt{purse}, \texttt{snowman}                                                                          & 2           \\
        \texttt{owl}                      & \texttt{mountain}                                                                                         & 1           \\
        \texttt{paintbrush}               & \texttt{moon}, \texttt{pillow}, \texttt{remote\_control}                                                  & 3           \\
        \texttt{palm\_tree}               & \texttt{sink}, \texttt{sun}                                                                               & 2           \\
        \texttt{paper\_clip}              & \texttt{penguin}, \texttt{rainbow}, \texttt{star}                                                         & 3           \\
        \texttt{penguin}                  & \texttt{paper\_clip}, \texttt{sink}, \texttt{sun}, \texttt{vase}                                          & 4           \\
        \texttt{piano}                    & \texttt{mug}, \texttt{strawberry}, \texttt{wine\_bottle}                                                  & 3           \\
        \texttt{picture\_frame}           & \texttt{umbrella}                                                                                         & 1           \\
        \texttt{pig}                      & \texttt{nose}, \texttt{train}                                                                             & 2           \\
        \texttt{pillow}                   & \texttt{paintbrush}, \texttt{sun}, \texttt{train}                                                         & 3           \\
        \texttt{pineapple}                & \texttt{paper\_clip}                                                                                      & 1           \\
        \texttt{pizza}                    & \texttt{mug}, \texttt{wristwatch}                                                                         & 2           \\
        \texttt{police\_car}              & \texttt{spoon}                                                                                            & 1           \\
        \texttt{purse}                    & \texttt{star}                                                                                             & 1           \\
        \texttt{rabbit}                   & \texttt{sandwich}, \texttt{sock}                                                                          & 2           \\
        \texttt{radio}                    & \texttt{light\_bulb}, \texttt{pineapple}, \texttt{zebra}                                                  & 3           \\
        \texttt{rainbow}                  & \texttt{mailbox}, \texttt{paper\_clip}, \texttt{snowman}                                                  & 3           \\
        \texttt{remote\_control}          & \texttt{paintbrush}, \texttt{sailboat}                                                                    & 2           \\
        \texttt{sailboat}                 & \texttt{moon}, \texttt{remote\_control}                                                                   & 2           \\
        \texttt{sandwich}                 & \texttt{rabbit}, \texttt{soccer\_ball}, \texttt{star}                                                     & 3           \\
        \texttt{school\_bus}              & \texttt{traffic\_light}                                                                                   & 1           \\
        \texttt{scissors}                 & \texttt{necklace}                                                                                         & 1           \\
        \texttt{shoe}                     & \texttt{nose}, \texttt{popsicle}                                                                          & 2           \\
        \texttt{sink}                     & \texttt{palm\_tree}, \texttt{penguin}, \texttt{suitcase}, \texttt{sun}, \texttt{television}               & 5           \\
        \texttt{skateboard}               & \texttt{mailbox}, \texttt{police\_car}                                                                    & 2           \\
        \texttt{smiley\_face}             & \texttt{soccer\_ball}, \texttt{suitcase}                                                                  & 2           \\
        \texttt{snowman}                  & \texttt{leaf}, \texttt{onion}                                                                             & 2           \\
        \texttt{soccer\_ball}             & \texttt{smiley\_face}, \texttt{television}                                                                & 2           \\
        \texttt{sock}                     & \texttt{rabbit}, \texttt{soccer\_ball}, \texttt{television}                                               & 3           \\
        \texttt{spoon}                    & \texttt{rainbow}                                                                                          & 1           \\
        \texttt{star}                     & \texttt{paper\_clip}, \texttt{sandwich}, \texttt{umbrella}                                                & 3           \\
        \texttt{strawberry}               & \texttt{light\_bulb}, \texttt{piano}, \texttt{table}, \texttt{tree}, \texttt{vase}, \texttt{wine\_bottle} & 6           \\
        \texttt{suitcase}                 & \texttt{rabbit}, \texttt{sink}                                                                            & 2           \\
        \midrule \textbf{Total}           & 42 unique left objects and 100 horizontal tasks                                                        & 100         \\
        \bottomrule
    \end{tabular}
\end{table*}

\begin{figure*}[t]
    \centering
    \includegraphics[width=\textwidth]{
        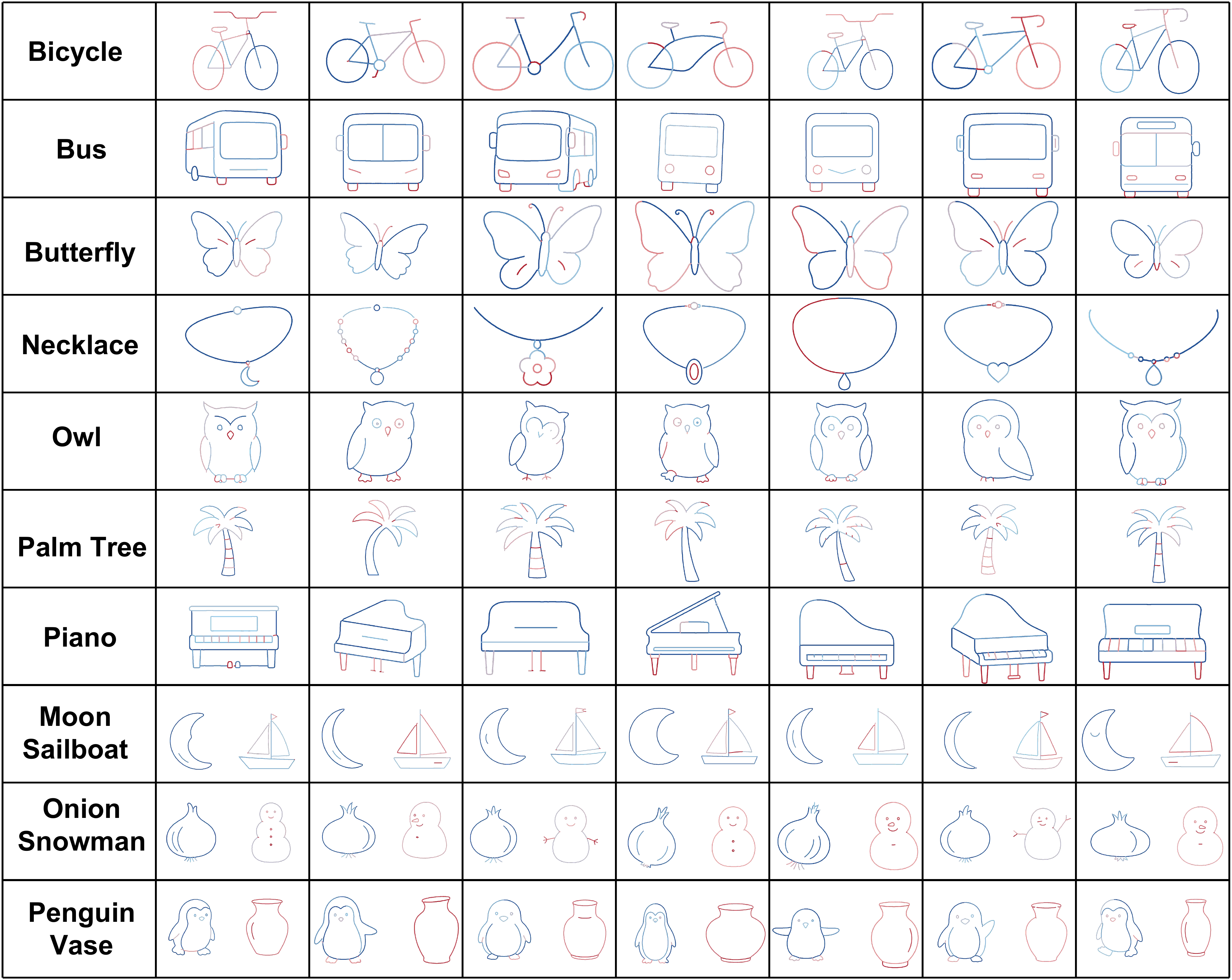
    }
    \caption{Stroke-order and geometry diversity in \myenv. Examples show different
    object shapes and drawing sequences, with colors indicating stroke order
    from dark blue to light blue, light red, and dark red.}
    \label{fig:app_display_dataset}

    \vspace{-10pt}
\end{figure*}
Figure~\ref{fig:app_display_dataset} visualizes 10 representative tasks from the
dataset. For each single-object class, the demonstrations exhibit variations in stroke
order and object geometry, reflecting the multimodal nature of sketch generation.
These single-object sketches can also be combined into compositional tasks that require
the agent to draw multiple objects according to language-specified spatial relations.

\begin{figure*}[ht]
   \centering
   \includegraphics[width=\textwidth]{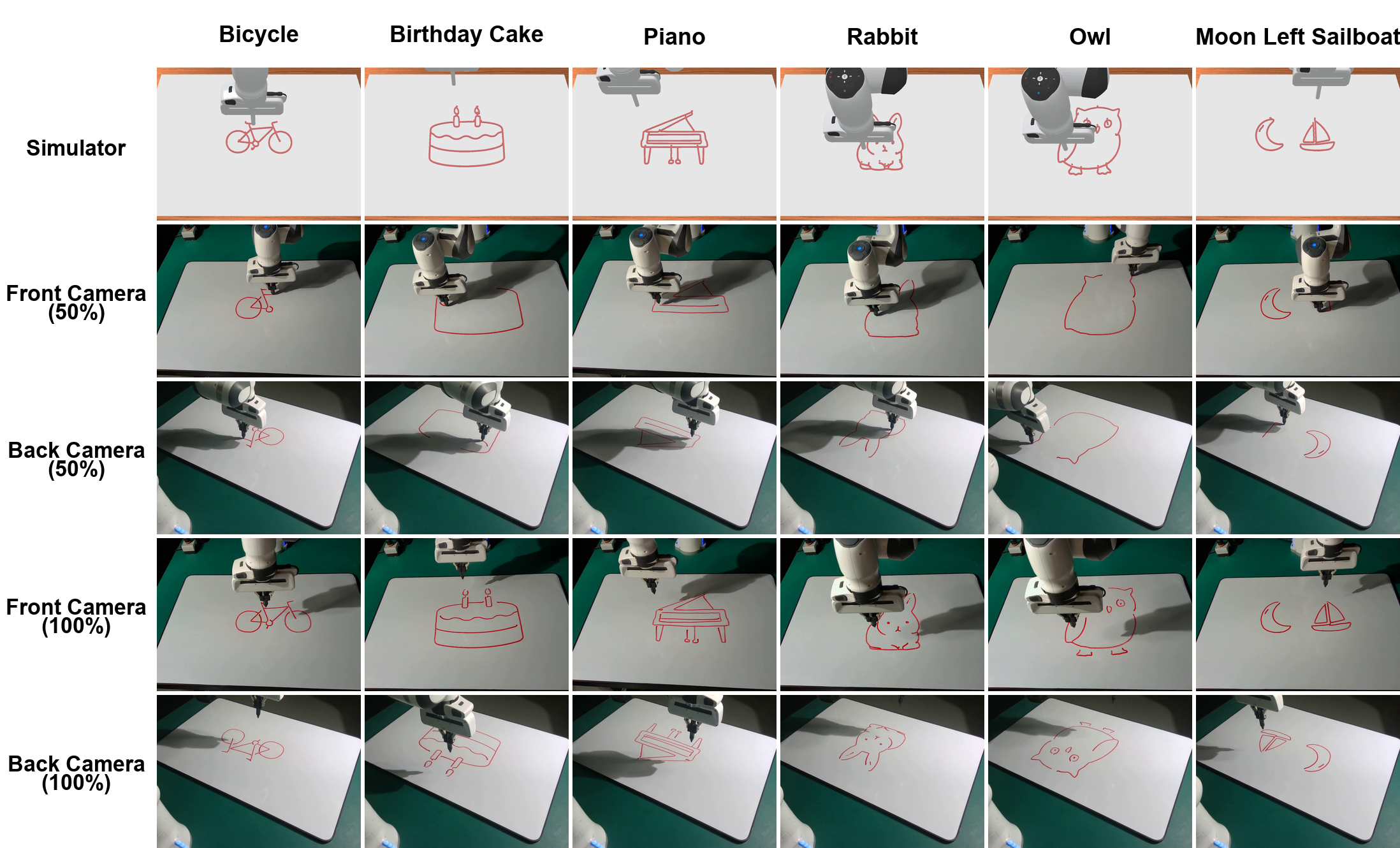}    
   \caption{
  Visualization of the real-robot dataset. The first row shows the corresponding simulated data. The second and third rows show intermediate stages of real-robot data collection. The last two rows show completed real-robot demonstrations.}
   \label{fig:app_real_robot_dataset}
\vspace{-10pt}
\end{figure*}

In addition, Figure~\ref{fig:app_real_robot_dataset} shows examples from our real-robot dataset, illustrating the correspondence between simulated trajectories and physical robot executions. 
To collect real-robot data, we replay the SVG trajectories generated by the pipeline described in Section~\ref{app:algorithms}.
This data-collection process follows a planned trajectory and therefore does not
require closed-loop visual feedback from the cameras. This open-loop replay procedure makes the real-robot dataset easy to construct: once SVG trajectories are available, additional demonstrations can be collected by directly replaying them on the robot without manual teleoperation or closed-loop policy execution.
As shown in Figure~\ref{fig:app_real_robot_dataset},
the robot can execute representative drawing tasks, such as \emph{birthday cake},
on the physical canvas. 
Each sample includes a language instruction, canvas observations from the front and rear cameras, the robot proprioceptive state, and the corresponding expert end-effector action. Either camera view can be used as the visual observation input.

\subsection{Comparison with Other Sketch Sources and LIBERO}
\label{app:dataset_comparison}

\begin{figure*}[ht]
    \vspace{-5pt}
    \centering
    \includegraphics[width=\textwidth]{         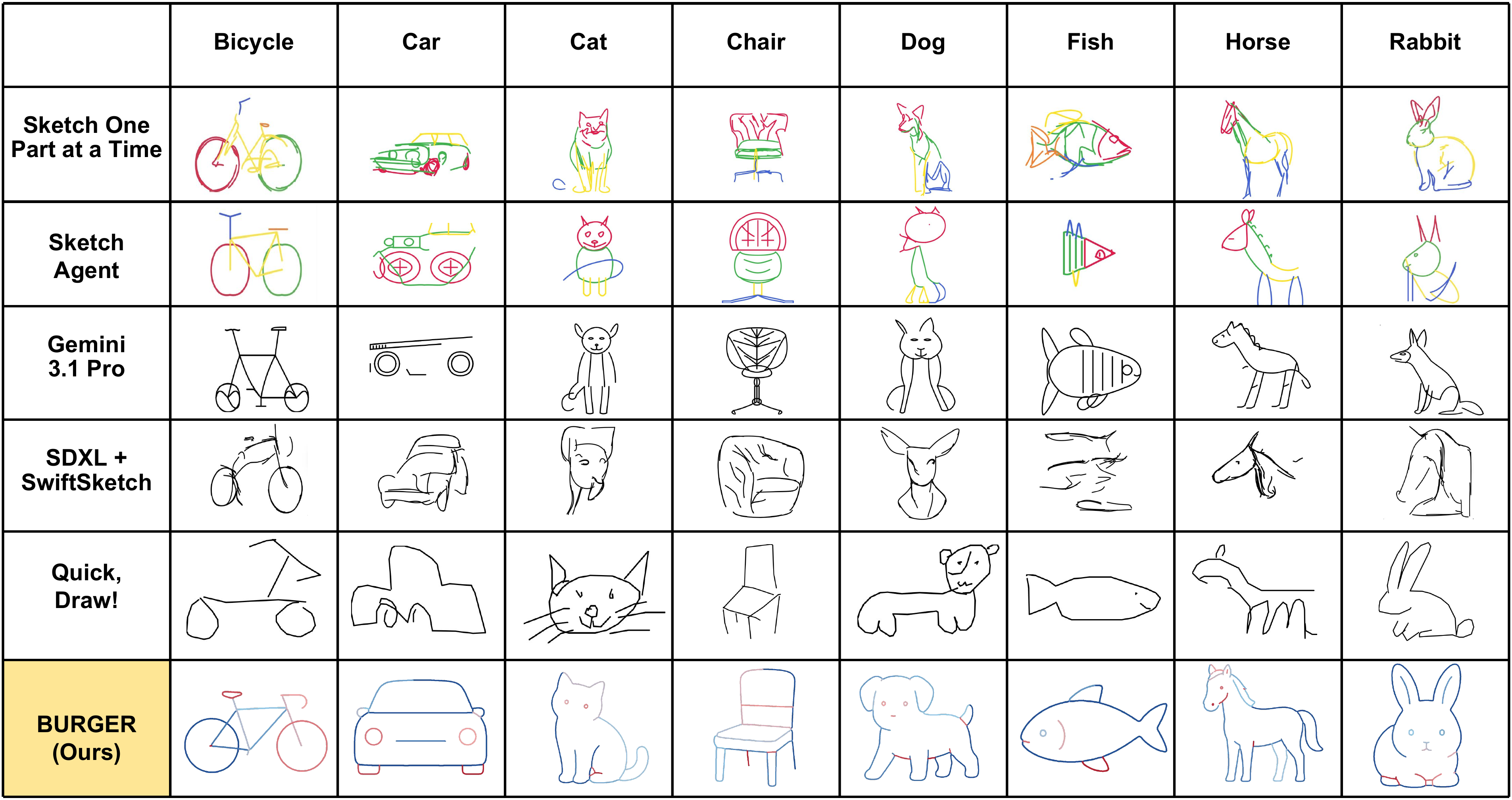
    }
    \caption[]{Data comparison with other sketch generation methods. }
    \label{fig:appendix_sketch_comparision}
\end{figure*}

In Figure~\ref{fig:appendix_sketch_comparision}, we compare our generated
sketches with representative sketch-generation methods and datasets, including Sketch
One Part at a Time~\cite{du2026sketch}, SketchAgent~\cite{vinker2025sketchagent},
Gemini 3.1 Pro~\cite{google2026gemini31pro}, SDXL~\cite{sdxl}+SwiftSketch~\cite{swiftsketch}, and Google Quick,
Draw!~\cite{david2017neural}. 
Compared with these alternatives, our generated sketches provide a favorable balance between visual recognizability and stroke simplicity: they contain a moderate number of lines, smooth contours, and clear object structure. These properties make them well suited for downstream conversion into robot-executable drawing trajectories. Moreover, because our pipeline is prompt-driven, it can scalably generate diverse sketch data conditioned on different object categories and language instructions.

We use a count-matched setting of
40 instructions and approximately 42 episodes per instruction. Each trajectory is
resampled to 32 points and represented by its start-relative $xy$ path and step
vectors. A shared $k$-means codebook is fitted on the pooled trajectory descriptors,
and the effective number of trajectory modes is defined as
\begin{equation}
N_{\mathrm{eff}}=\exp\left(-\sum_{j=1}^{k}p_j\log p_j\right),
\end{equation}
where $p_j$ denotes the fraction of trajectories assigned to cluster $j$. 

\begin{wrapfigure}{r}{0.3\textwidth}
    \centering
    \includegraphics[width=\linewidth]{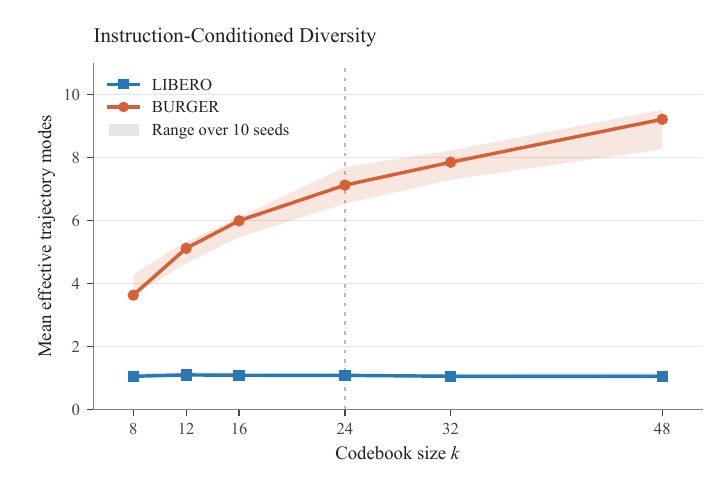}
    \caption{Codebook-size sensitivity. Solid curves show the reference seed; shaded bands show the min–max across ten initialization seeds on a fixed trajectory sample.}
    \label{fig:burger-clustering-sensitivity}
    \vspace{-10pt}
\end{wrapfigure}
We resample each trajectory to 32 uniformly spaced points in normalized
time and concatenate its start-relative planar positions and successive
displacements into a 126-dimensional descriptor. After standardization
using pooled statistics, we fit a shared $k$-means codebook with
$k$-means++ initialization. Cluster frequencies are computed within each
instruction to obtain $N_{\mathrm{eff}}(\ell;k)$, then averaged equally
over instructions. Figure~\ref{fig:burger-clustering-sensitivity} shows
the results for $k\in\{8,12,16,24,32,48\}$ with all other settings fixed.
The reference curves show LIBERO remaining at 1.05--1.10 effective modes
per instruction, while BURGER increases from 3.63 to 9.22, with a
6.57-fold difference at $k=24$. BURGER consistently exceeds LIBERO across
ten random initializations, supporting the robustness of its greater
instruction-conditioned trajectory diversity across the tested
clustering resolutions.

\section{Experimental Details}

\subsection{VLA Post-Training Protocol}
\label{app:vla_postraining}

Our dataset is stored in the LeRobot format~\cite{cadene2024lerobot}. Each training sample consists of a language instruction, the current canvas observation, the robot proprioceptive state, and the corresponding expert action. The language instruction follows the task prompt used during data generation, e.g., ``draw an apple'' for single-object tasks and ``draw an apple to the left of a cup'' for compositional tasks. For visual input, we use the rendered canvas observation from the simulator. The action target is the robot end-effector command recorded during SVG replay, as described in the SVG-to-trajectory pipeline in the main paper.

We keep the post-training protocol consistent across models whenever possible. For \(\pi_0\) and \(\pi_{0.5}\), we fine-tune the continuous-action policy using their original flow-matching action objective. For \(\pi_0\)-FAST, actions are first converted into FAST tokens and the model is trained with the autoregressive action-token prediction objective. We use a batch size of $32$ for all models. All models are evaluated without additional task-specific adaptation after post-training.

During evaluation, the policy receives the language prompt, the current canvas observation, and the robot proprioceptive state, and predicts an end-effector command or action chunk. An episode terminates when the robot end-effector is lifted above a predefined height threshold; therefore, the policy must learn both the drawing behavior and the appropriate termination behavior.
We use the same initial-state distribution and maximum episode length across models to ensure a fair comparison. For single-object evaluation, we use 10 object classes: alarm clock, apple, backpack, banana, bathtub, bench, bicycle, birthday cake, blueberry, and book. For compositional evaluation, we test on unseen spatial compositions. For each task, we run 10 evaluation episodes and report the average score computed by our evaluation metric. The final canvas is used for visual evaluation, while the full trajectory is used for analyzing failure modes and real-robot transfer.

\subsection{Rollout Evaluation}
\label{app:evaluation_details}

\textbf{Latent-based Similarity Metric.} We construct a sketch-specific metric space on top
of a frozen DINOv3 backbone. Given a rendered canvas image $I$, DINOv3 produces a
CLS token, which provides a global image representation, and a sequence of patch
embeddings, which preserve local visual details. We concatenate the CLS token with
the averaged patch embeddings and pass the resulting feature through a lightweight
two-layer MLP projection head with output dimension $256$, followed by
$\ell_{2}$ normalization. This yields a normalized sketch embedding $z(I)\in\mathbb{R}
^{256}$.

The embedding $z(I)$ is then passed to a cosine classifier. For class $c$, the classifier
maintains a learnable class weight $w_{c}$ and computes the logit as a scaled cosine
similarity:
\[
    \ell_{i,c}= \gamma z_{i}^{\top}\frac{w_{c}}{\|w_{c}\|_{2}},
\]
where $z_{i}=z(I_{i})$ and $\gamma$ is a learnable logit scale.

We train the projection head and cosine classifier using a combination of cross-entropy
classification and supervised contrastive learning:

\begin{equation}
\mathcal{L}
=
\mathcal{L}_{\mathrm{CE}}
+
\lambda \mathcal{L}_{\mathrm{SupCon}},
\end{equation}

where

\begin{align}
\mathcal{L}_{\mathrm{CE}}
&=
-\frac{1}{B}\sum_{i=1}^{B}
\log
\frac{\exp(\ell_{i,y_i})}
{\sum_{c=1}^{C}\exp(\ell_{i,c})},
\\
\mathcal{L}_{\mathrm{SupCon}}
&=
-\frac{1}{|\mathcal{A}|}
\sum_{i\in\mathcal{A}}
\frac{1}{|P(i)|}
\sum_{p\in P(i)}
\log
\frac{\exp(z_i^\top z_p/\tau)}
{\sum_{a\neq i}\exp(z_i^\top z_a/\tau)}.
\end{align}

Here, $B$ is the batch size, $C$ is the number of object classes, $y_{i}$ is the
ground-truth class label for image $I_{i}$, $\tau$ is the contrastive temperature,
$P(i)$ denotes the set of same-class positives for anchor $i$, and $\mathcal{A}=\{
i:|P(i)|>0\}$ is the set of valid anchors. The cross-entropy term makes the
embedding predictive of the object category, while the supervised contrastive term
encourages drawings from the same class to cluster together and drawings from different
classes to separate in the learned metric space. During training, the DINOv3 backbone
is kept frozen, and only the projection head and cosine classifier are optimized.

After training, we compute one class center for each object class using the training
split. For class $c$ with training drawings $\mathcal{D}^{\mathrm{train}}_{c}$, its
class center is
\[
    q_{c}= \frac{ \frac{1}{|\mathcal{D}^{\mathrm{train}}_c|}\sum_{I_i\in\mathcal{D}^{\mathrm{train}}_c}z(I_{i})
    }{ \left\| \frac{1}{|\mathcal{D}^{\mathrm{train}}_c|}\sum_{I_i\in\mathcal{D}^{\mathrm{train}}_c}z(I_{i})
    \right\|_{2}}.
\]
For a generated drawing with target class $y$, we compare its normalized
embedding with the corresponding class center:
\[
    s(I, y)=z(I)^{\top}q_{y}.
\]
We use this target-class similarity as the evaluation signal. A higher
similarity indicates that the generated drawing is closer to the target object
class in the learned sketch embedding space.

For single-object tasks, we score the final canvas image by its similarity to the
class center of the prompted object (i.e., $s(I, y)$). For compositional tasks,
we first split the foreground strokes into object regions using a vertical or horizontal separator with the lowest foreground-pixel density, evaluate each region against its corresponding object
class center, and average the two similarities. We further penalize predictions whose
detected ordering does not match the prompted spatial relation.

\textbf{Correlation with Human Judgments.}
We conduct a user study on 10 single-object classes to evaluate whether our latent-based similarity metric
is aligned with human judgments. For each class, 66 participants are shown five drawings:
four model-generated drawings produced by two VLA models and one ground-truth
drawing from our dataset. Participants are asked to rank the five drawings from best
to worst according to how well each drawing depicts the prompted object class.
We then evaluate the same drawings using our DINOv3-based~\cite{simeoni2025dinov3} similarity metric and CLIPScore~\cite{hessel2021clipscore}.

To quantify alignment with human judgments, we compute three complementary measures.
First, we use Spearman's rank correlation coefficient and Kendall $\tau_{b}$ correlation coefficient to compare the aggregate human ranking with rankings induced by each metric. For each drawing, the aggregate human score is computed as the negative mean human rank, so that larger values correspond to better human preference. Kendall's $\tau_{b}$ is used because it accounts for tied ranks, which occur frequently for CLIPScore due to score saturation. Second, we compute pairwise agreement by comparing each participant's preference over every within-class pair of drawings with the corresponding pairwise preference induced by the metric. Thus, Spearman's $\rho$ and Kendall's $\tau_{b}$ evaluate agreement with the aggregate human ranking, while pairwise agreement measures consistency with individual human judgments.

As shown in Table~\ref{tab:correlation}, our latent-based similarity metric achieves
stronger correlation with human judgments than CLIPScore across all three
measures. Specifically, our metric achieves a Spearman correlation of $0.834$,
Kendall's $\tau_{b}$ of $0.655$, and pairwise agreement of $0.875$, outperforming
CLIPScore, which obtains $0.592$, $0.433$, and $0.803$, respectively. These
results suggest that the proposed latent-based similarity metric  provides
a practical and human-aligned proxy for the evaluation of language-conditioned drawing policies.

\begin{table}[th]
    \caption{Correlation between metrics and human judgments.}
    \label{tab:correlation}
    \centering
    \small
    \setlength{\tabcolsep}{4pt}
    \begin{tabularx}
        {\columnwidth}{ >{\centering\arraybackslash}X >{\centering\arraybackslash}X >{\centering\arraybackslash}X >{\centering\arraybackslash}X >{\centering\arraybackslash}X}
        \toprule Metric & Spearman $\rho$ & Kendall $\tau_{b}$ & Pairwise
        Agreement \\
        \midrule CLIPScore~\cite{hessel2021clipscore} & 0.592 & 0.433 & 0.803 \\
        \midrule \textbf{Similarity metric} & 0.834 & 0.655 & 0.875 \\
        \bottomrule
    \end{tabularx}
    \vspace{-15pt}
\end{table}

\subsection{Real-Robot Implementation}
\label{app:real_robot}

\textbf{Inference.} For real-robot inference, we first align the real camera
observations with the simulated canvas coordinate system. Our setup uses two Intel
RealSense cameras, placed at the front and back of the workspace, to reduce occlusion
caused by the robot arm. 
To extract the drawing strokes on the canvas, we use depth observations to segment out the robot arm and then estimate the visible stroke regions from each camera view. The stroke observations from the two views are merged and provided as visual input to the trained VLA policy. The policy outputs target TCP positions, which are executed by the Franka Panda robot arm.

\textbf{Human--Robot Interaction.}
Our perception pipeline also supports human--robot collaborative drawing. During execution, depth observations are used to detect when a human hand enters the canvas region. When a hand is detected, policy inference is temporarily paused while the user modifies the drawing, either by adding new strokes or erasing existing ones. After the user leaves the workspace, the updated canvas is captured and processed using the same stroke-extraction and multi-view merging pipeline described above. The resulting canvas representation is then provided to the VLA policy as a new observation, and the robot resumes drawing conditioned on the updated canvas state. This procedure enables iterative collaboration between a human user and the robot within a shared drawing workspace.

\section{Additional Results}

\subsection{Additional Scaling Results}
\label{app:additional_scaling}

\begin{figure*}[tb]
    \centering

    \begin{subfigure}{\textwidth}
        \centering
        \includegraphics[width=\linewidth]{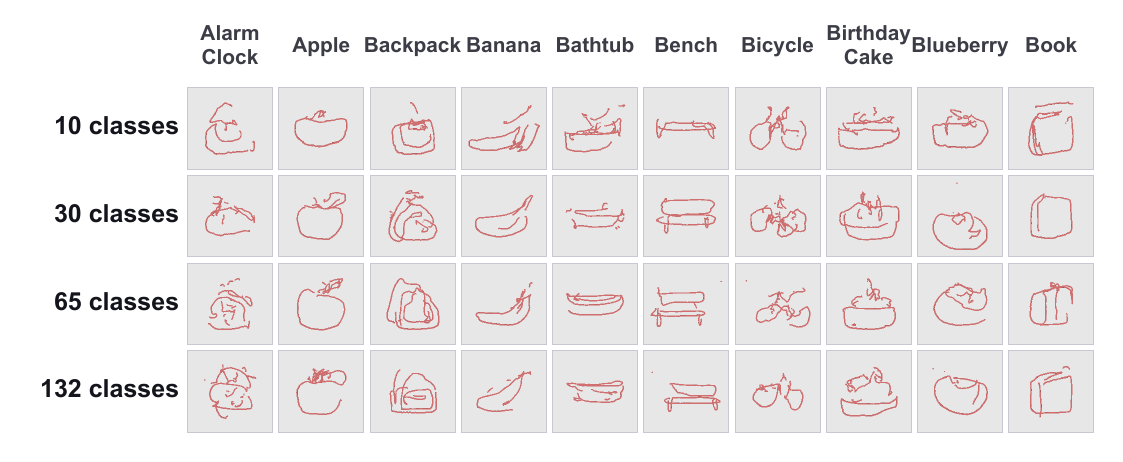}
        \caption{10 demonstrations per class.}
        \label{fig:pi05_matrix_10demos_4x10}
    \end{subfigure}

    \vspace{0.5em}

    \begin{subfigure}{\textwidth}
        \centering
        \includegraphics[width=\linewidth]{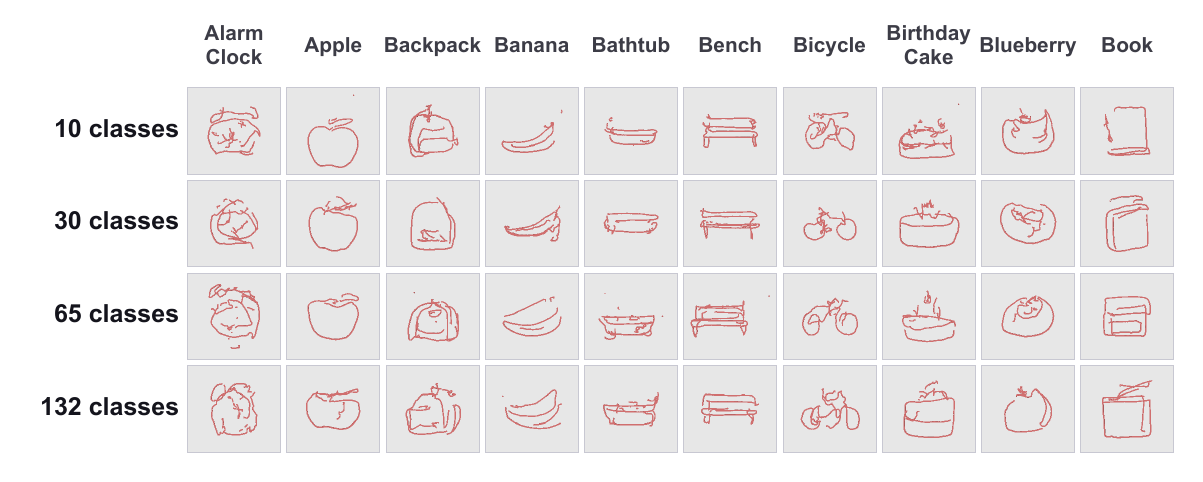}
        \caption{50 demonstrations per class.}
        \label{fig:pi05_matrix_50demos_4x10}
    \end{subfigure}

    \vspace{0.5em}

    \begin{subfigure}{\textwidth}
        \centering
        \includegraphics[width=\linewidth]{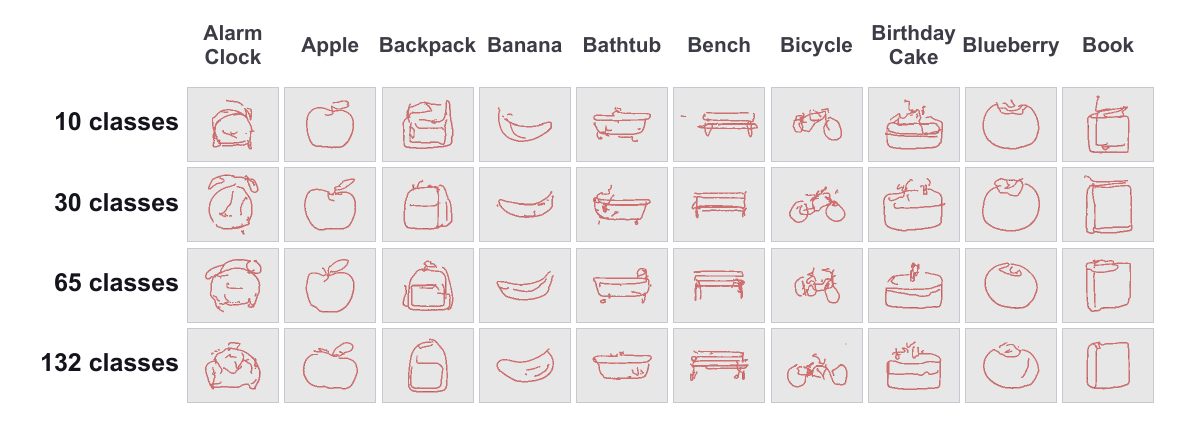}
        \caption{100 demonstrations per class.}
        \label{fig:pi05_matrix_100demos_4x10}
    \end{subfigure}

    \caption{
    Qualitative data-scaling results for \(\pi_{0.5}\). Panels (a), (b), and (c) show models trained with 10, 50, and 100 demonstrations per class, respectively. Within each panel, rows correspond to different numbers of training classes. Increasing either the number of demonstrations per class or the number of training classes improves drawing quality.
    }
    \label{fig:pi05_scaling_4x10}
\end{figure*}

Figure~\ref{fig:pi05_scaling_4x10} provides additional qualitative results for \(\pi_{0.5}\) under different data scales, varying both the number of demonstrations per class and the number of training classes. Consistent with the quantitative results in Figure~\ref{fig:scaling}, performance improves when either the number of demonstrations or the number of classes increases. We also observe that increasing the number of demonstrations per class has a particularly strong effect on drawing quality, suggesting that sufficient within-class coverage is important for learning reliable stroke-level action generation.
\begin{figure*}[tb]
   \centering
   \includegraphics[width=\textwidth]{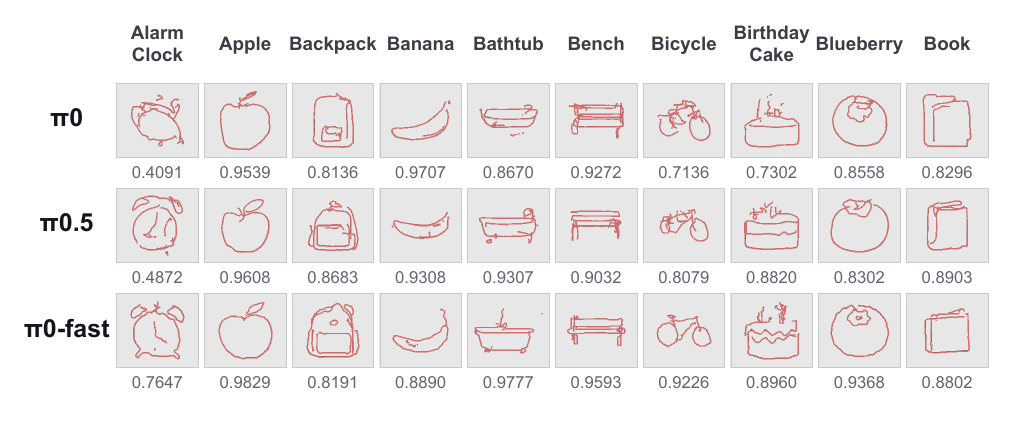}    
   \caption{
   Qualitative comparison of representative VLA models on \myenv. }
   \label{fig:three_model_compare}

\vspace{-10pt}
\end{figure*}

We further compare representative rollouts from different VLA models in Figure~\ref{fig:three_model_compare}. 
For each model and task, we select the rollout with the highest score according to our evaluation metric.

\subsection{Additional Multimodal Rollouts}
\label{app:additional_multimodal}

\begin{figure*}[tb]
   \centering
        \includegraphics[width=\linewidth]{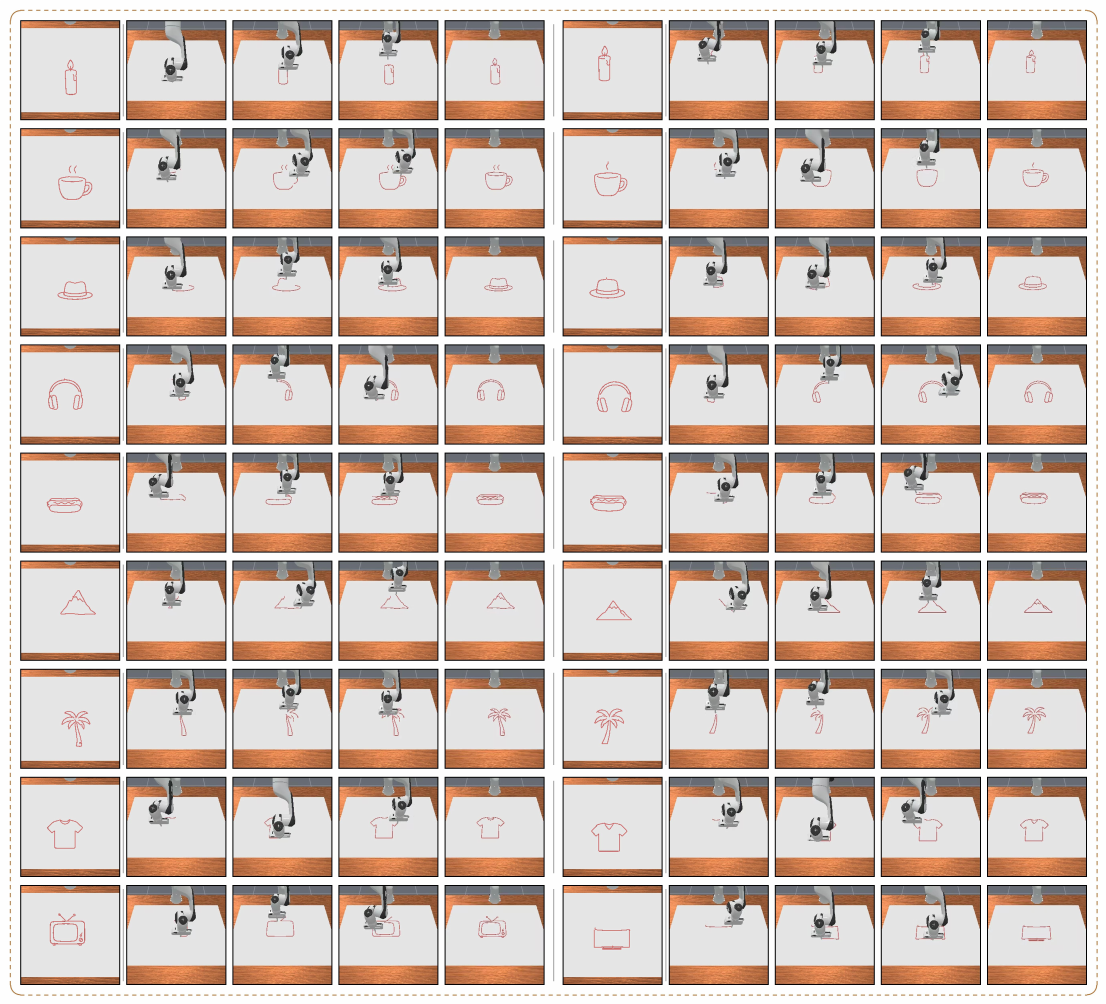}
        \caption{Multimodal drawing behaviors in the training set. The same object class can be represented by different stroke orders, shapes, and local geometric variations.}
        \label{fig:app_multi_modality_trainset}

\vspace{-10pt}
\end{figure*}

\begin{figure*}[tb]
        \centering
        \includegraphics[width=\linewidth]{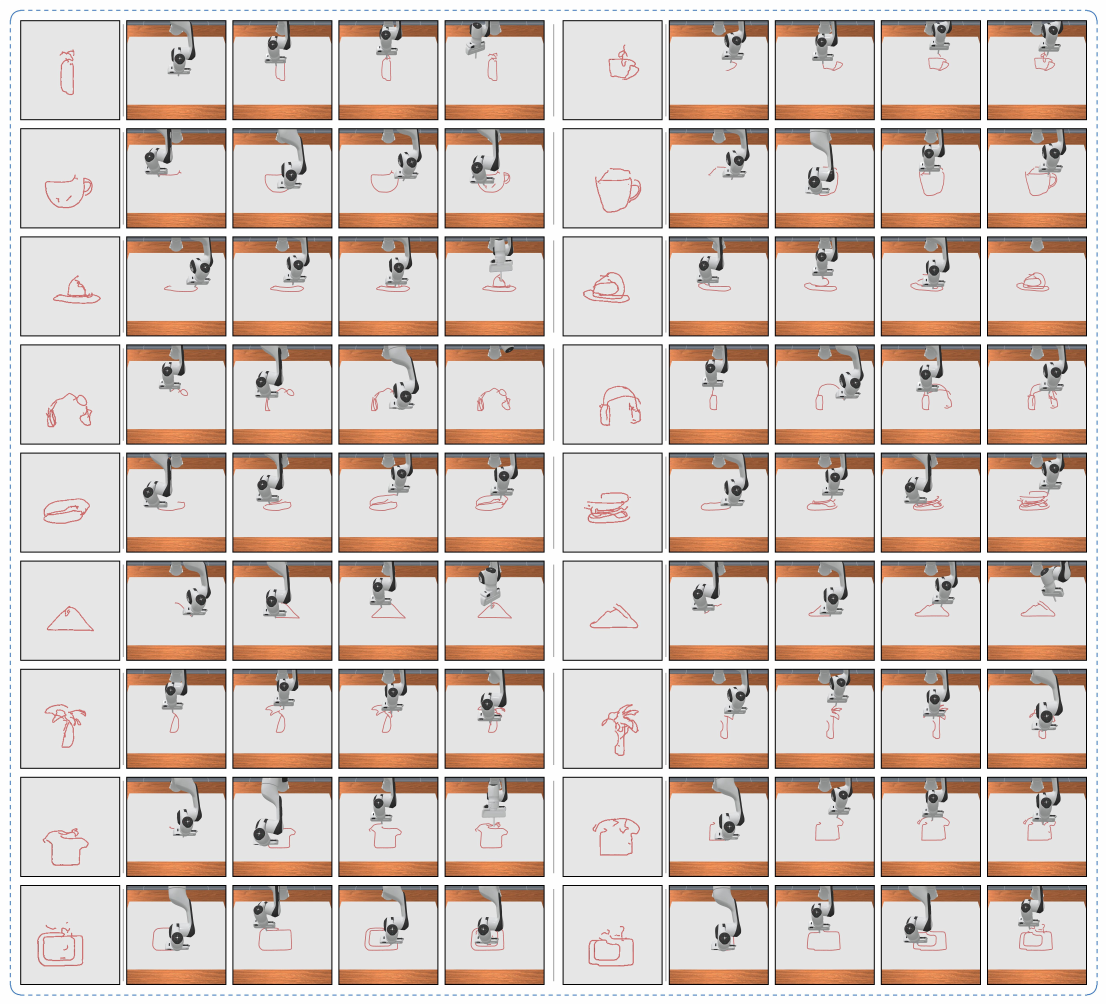}
        \caption{Multimodal drawing behaviors in VLA rollouts. The generated drawings show that trained VLA models can produce multiple valid realizations for the same language instruction.}
        \label{fig:app_multi_modality_evalset}
      \vspace{-15pt}  
\end{figure*}

Figure \ref{fig:app_multi_modality_trainset} and Figure \ref{fig:app_multi_modality_evalset} provide additional examples of multimodal drawing behaviors in \myenv. The training demonstrations show that the same object can be drawn with different stroke orders and geometric variations, while the model rollouts show that VLA policies can reproduce multiple valid drawing modes after training.

\subsection{Additional Spatial-Composition Results}
\label{additional_composition}

\begin{figure*}[tb]
   \centering
        \includegraphics[width=\linewidth]{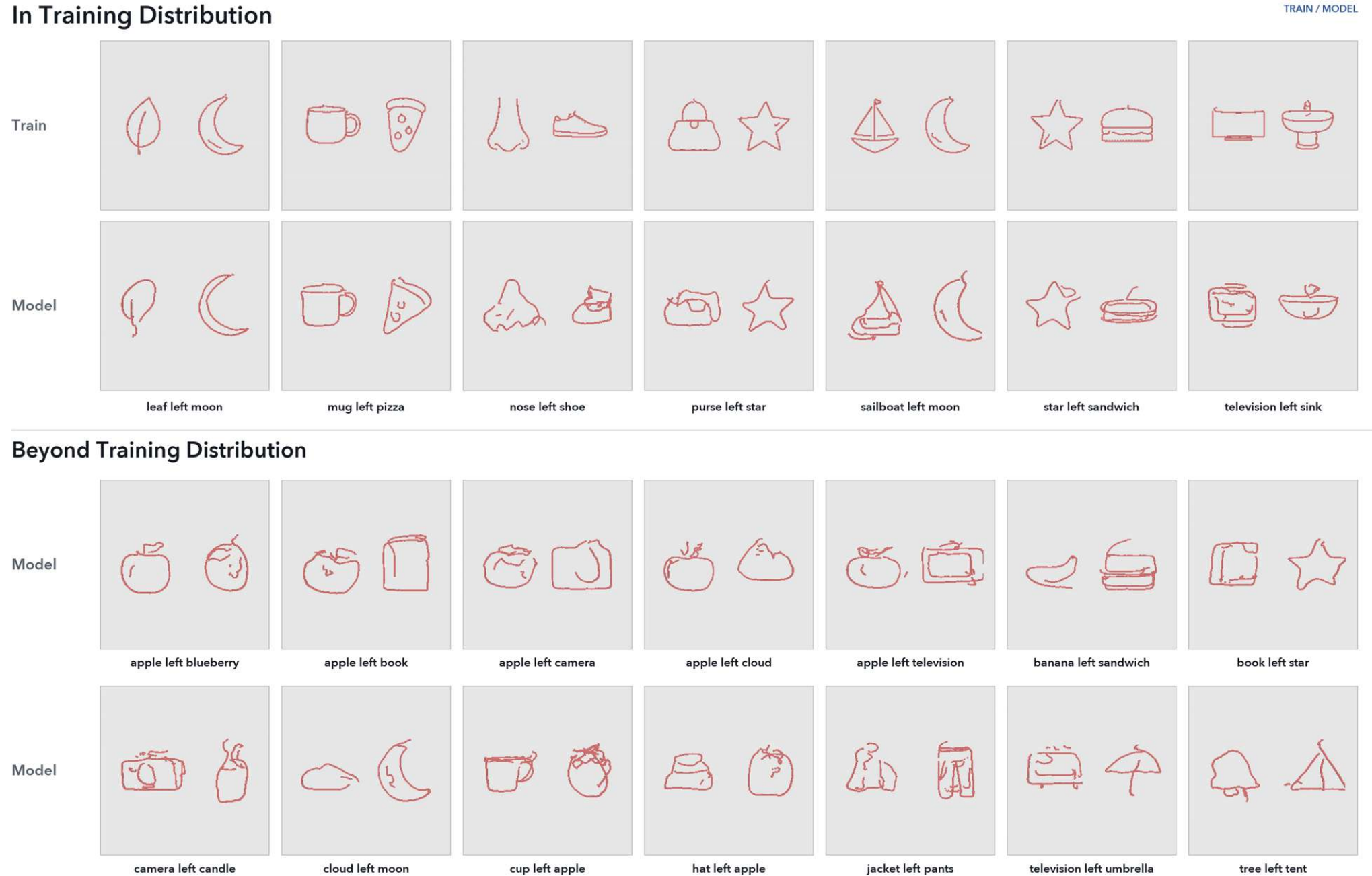}
        \caption{Above: In-distribution spatial-relation grounding. VLA models learn to generate compositional drawings for object pairs observed during training. Below: Generalization to unseen object compositions. VLA models can apply learned spatial relations to object pairs that are not observed during training.}
        \label{fig:compositional_tasks}
\vspace{-10pt}
\end{figure*}

We further provide additional qualitative examples of spatial-relation learning in Figure~\ref{fig:compositional_tasks}.  Figure~\ref{fig:compositional_tasks} shows model rollouts on in-distribution compositional tasks, where the object composition is observed during training, as well as rollouts on unseen object compositions, demonstrating that VLA models can apply learned spatial relations to new object pairs.

\end{document}